\documentclass[letterpaper, 10pt, conference]{ieeeconf}      

\IEEEoverridecommandlockouts                              

\usepackage{color}
\usepackage{graphicx}
\usepackage{amsmath}
\usepackage{amssymb}
\usepackage{tabularx}
\usepackage{subcaption}
\usepackage{url}
\usepackage{cite}
\usepackage{algorithm} 
\usepackage{algorithmic}
\usepackage{blkarray}

\DeclareMathOperator*{\argmax}{\arg\!\max}

\title{\LARGE \bf  3D Deformable Object Manipulation using Fast Online Gaussian Process Regression} 

\author{Zhe Hu \and Peigen Sun \and Jia Pan%
\thanks{The authors are with the Department of Mechanical and Biomedical Engineering, the City University of Hong Kong, Hong Kong.}%
}

\begin{document}
\maketitle

\begin{abstract}
In this paper, we present a general approach to automatically visual-servo control the position and shape of a deformable object whose deformation parameters are unknown. The servo-control is achieved by online learning a model mapping between the robotic end-effector's movement and the object's deformation measurement. The model is learned using the Gaussian Process Regression (GPR) to deal with its highly nonlinear property, and once learned, the model is used for predicting the required control at each time step. To overcome GPR's high computational cost while dealing with long manipulation sequences, we implement a fast online GPR by selectively removing uninformative observation data from the regression process. We validate the performance of our controller on a set of deformable object manipulation tasks and demonstrate that our method can achieve effective and accurate servo-control for general deformable objects with a wide variety of goal settings. Experiment videos are available at 
\url{https://sites.google.com/view/mso-fogpr}.

\end{abstract}



\section{Introduction}

Manipulation of deformable objects is a challenging problem in robotic manipulation and has many important applications, including cloth folding~\cite{Miller:2011:AGA, Towner:2011:BCI},
string insertion~\cite{Wang:2015:AOM}, sheet tension~\cite{Kruse:2015:CHR}, robot-assisted surgery~\cite{Patil:2010:TAT}, and suturing~\cite{Schulman:2013:GIR}. 

Most previous work on robotic manipulation can be classified into two categories: some approaches did not explicitly model the deformation parameters of the object, and used vision or learning methods to accomplish tasks~\cite{Miller:2011:AGA,Twardon:2015:ISC,Schulman:2013:GIR,Hadfield:2015:BLW,Tang:2016:RMD,Yang:2017:RFT,Li:2015:FDO,Li:2015:RUG}. These methods focus on high-level policies but lack the capability to achieve accurate operation -- actually most of them are open-loop methods. Other approaches require a model about the object's deformation properties in terms of stiffness, Young' modules, or FEM coefficients, to design a control policy~\cite{McConachie:2016:BBM, Patil:2010:TAT, Bodenhagen:2014:AAR,Alarcon:2013:MFV,Alarcon:2016:ADM,Langsfeld:2016:OLP}. However, such deformation parameters are difficult to be estimated accurately and may even change during the manipulation process, especially for objects made by nonlinear elastic or plastic materials. These challenges leave the automatic manipulation of deformable objects an open research problem in robotics~\cite{Henrich:2012:RMO}. 

In this paper, we focus on designing servo-manipulation algorithm which can learn a nonlinear deformation function along with the manipulation process. The deformation function is efficiently learned using a novel online Gaussian process regression and is able to model the relation between the movement of the robotic end-effectors and the soft object's deformation adaptively during the manipulation. In this way, we design a nonlinear feedback controller that makes a good balance between exploitation and exploration and provides better convergence and dynamic behavior than previous work using linear deformation models such as ~\cite{Alarcon:2013:MFV,Alarcon:2016:ADM}. Our manipulation system successfully and efficiently accomplishes a set of different manipulation tasks for a wide variety of objects with different deformation properties.


\begin{figure}[t] 
\centering
\includegraphics[width=0.9\linewidth]{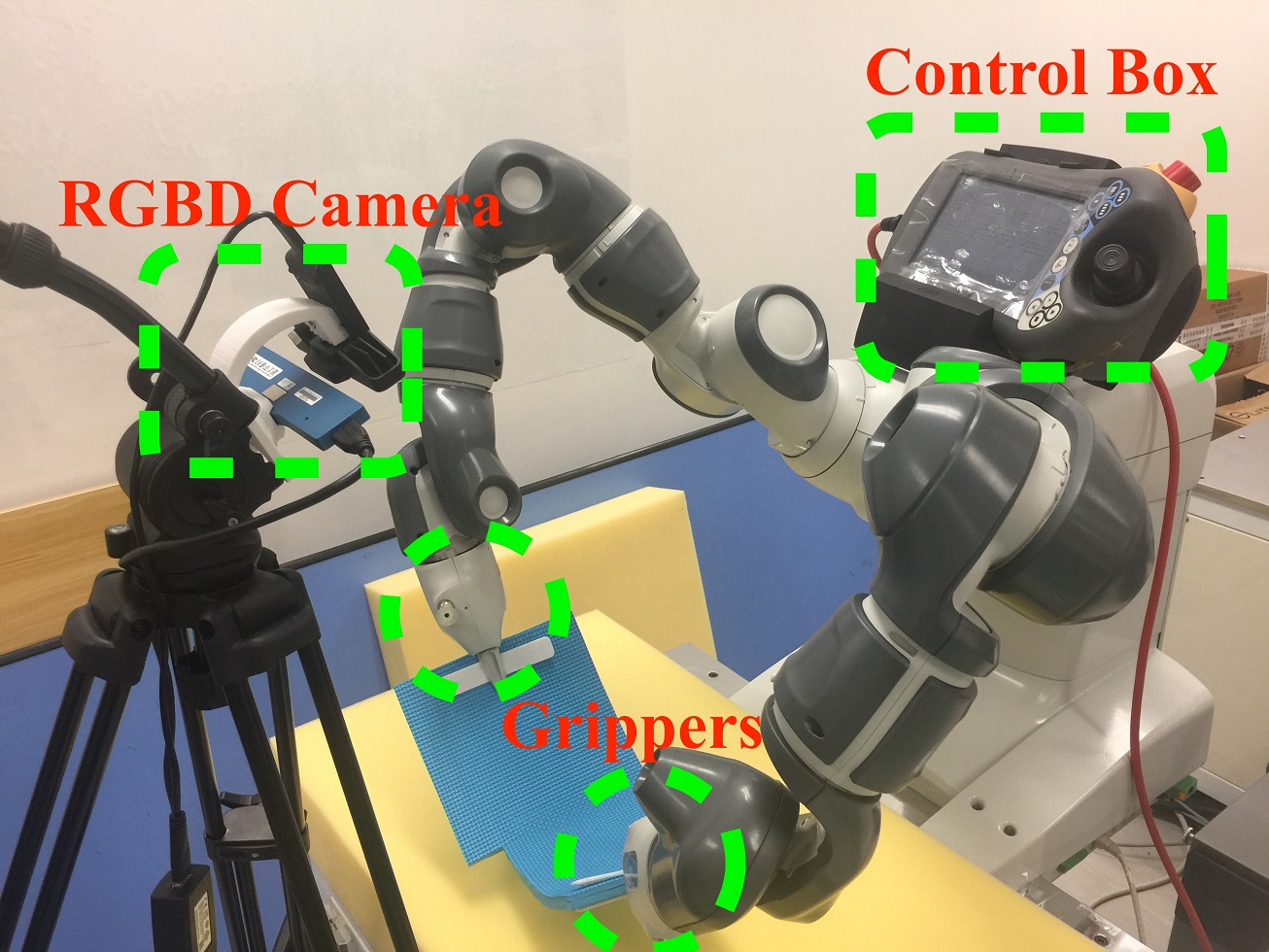}
\caption{Our robotic system for deformable object manipulation is made by two 3D cameras and one dual-arm ABB robot. }
\label{fig:manipulation-system}
\vspace*{-0.2in}
\end{figure}

\section{Related Work}
\label{sec:related}

Many robotic manipulation method for deformable objects have been proposed recent years. Early work~\cite{Saha:2007:MPD,Moll:2006:PPD} used knot theory or energy theory to plan the manipulation trajectories for linear deformable objects like ropes. Some recent work~\cite{Bai:2016:DMC} further considered manipulating cloths using dexterous cloths. These work required a complete and accurate knowledge about the object' geometric and deformation parameters and thus are not applicable in practice. 

More practical work used sensors to guide the manipulation process. ~\cite{Matsuno:2006:MDL} used image to estimate the knot configuration. ~\cite{Miller:2011:AGA} used vision to estimate the configuration of a cloth and then leverage gravity to accomplish folding tasks~\cite{Bell:2010:GNS}. ~\cite{Twardon:2015:ISC} used RGBD camera to identify the boundary components in clothes. ~\cite{Li:2015:FDO,Li:2015:RUG} first used vision to determine the status of the cloth, then optimized a set of grasp points to unfold the clothes on the table, and finally found a sequence of folding actions.  
Schulman \emph{et al.}~\cite{Schulman:2013:GIR} enabled a robot to accomplish complex multi-step deformation object manipulation strategies by learning from a set of manipulation sequences with depth images to encode the task status. Such learning from demonstration technique has further been extended using reinforcement learning~\cite{Hadfield:2015:BLW} and tangent space mapping~\cite{Tang:2016:RMD}. A deep learning-based end-to-end framework has also been proposed recently~\cite{Yang:2017:RFT}. A complete pipeline for clothes folding task including vision-based garments grasping, clothes classification and unfolding, model matching and folding has been described in~\cite{Doumanoglou:2016:FCA}. 

Above methods generally did not explicitly model the deformation parameters of the deformation objects, which is necessary for high-quality manipulation control. Some methods used uncertainty model~\cite{McConachie:2016:BBM} or heuristics~\cite{Berenson:2013:MDO, Wang:2015:AOM} to take into account rough deformation models during the manipulation process. Some work required an offline procedure to estimate the deformation parameters~\cite{Bodenhagen:2014:AAR}. There are several recent work estimating the object's deformation parameters in an online manner and then design controller accordingly. Navarro-Alarcon \emph{et al.}~\cite{Alarcon:2013:MFV,Alarcon:2016:ADM} used an adaptive and model-free linear controller to servo-control soft objects, where the object's deformation is modeled using a spring model~\cite{Hirai:2000:ISP}. \cite{Langsfeld:2016:OLP} learned the models of the part deformation depending on the end-effector force and grasping parameters in an online manner to accomplish high-quality cleaning task. A more complete survey about deformable object manipulation in industry is available in\cite{Henrich:2012:RMO}.

In this paper, we are using Gaussian process regression (GPR) to model and learn the deformation parameters of a soft object. Our method is motivated by several recent work focus on reducing the computational cost of the offline GPR. ~\cite{Snelson:2006:SGP} presented a sparse GPR method by selecting $M$ pseudo-input points from the $N$ training data to balance the computational cost and the model accuracy, where $M \ll N$. ~\cite{Rasmussen:2002:MGP,Snelson:2007:GPA} divided the input space of the Gaussian process model into smaller subspaces, and fit a local GPR for each subspace. ~\cite{Duy:2009:LGP} used many local GPRs and updated local models iteratively, in order to reduce the training time. Our method proposed an online sparse method for efficient deformation model adaptation during the manipulation process.


\section{Overview and Problem Formulation}
\label{sec:prob}
\begin{figure}[t] 
\centering
\includegraphics[width=0.9\linewidth]{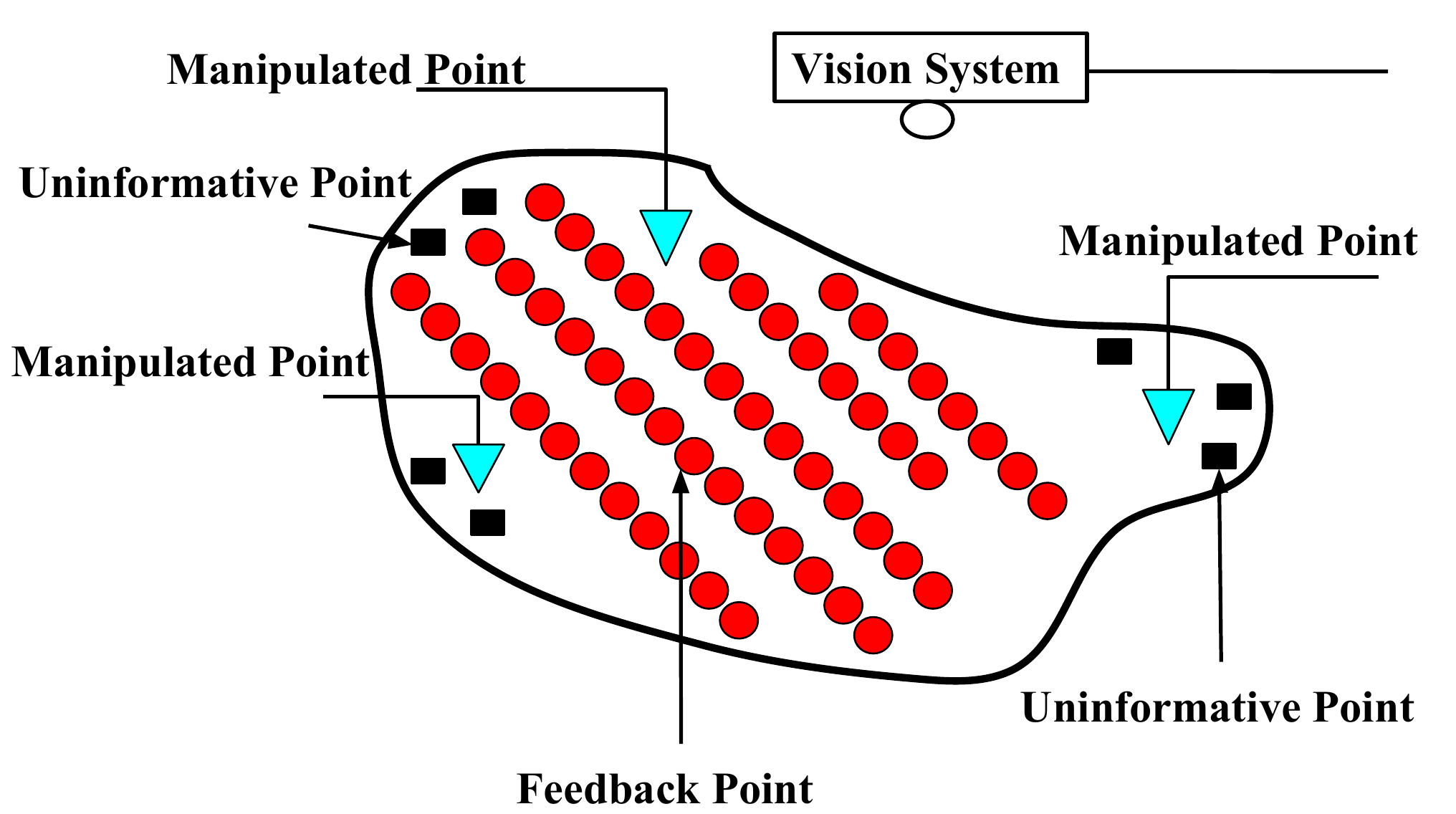}
\caption{We model a soft object using three classes of points： manipulated points, feedback points, and uninformative points.}
\label{fig:problem-formulation}
\vspace*{-0.2in}
\end{figure}

The problem of 3D deformable object manipulation can be formulated as follows. 
Similar to~\cite{Hirai:2000:ISP}, we describe an object as a set of discrete points, which are classified into three types: the manipulated points, the feedback points, and the uninformative points, as shown in Figure~\ref{fig:problem-formulation}. The manipulated points correspond to the positions on the object that are grabbed by the robot and thus is fixed relative to the robotic end-effectors; the feedback points correspond to the object surface regions that define the task goal setting and involve in the visual feedbacks; and the uninformative points correspond to other regions on the object. Given this setup, the deformable object manipulation problem is about how to move the manipulated points in order to drive the feedback points toward a required target configuration. 

Since the manipulation of deformable object is usually executed at a low speed to avoid vibration, we can reasonably assume that the object always lies in the quasi-static state where the internal forces caused by the elasticity of the object is balanced with the external force applied by the end-effector on the manipulated points. We use a potential energy function $U(\mathbf p^m, \mathbf p^f, \mathbf p^u)$ to formulate the elasticity of the deformable object, where the potential energy depends on all the points on the object, and vectors $\mathbf p^m$, $\mathbf p^f$ and $\mathbf p^u$ represent the stacked coordinates of all manipulated points, feedback points and uninformed points, respectively. The equation of equilibrium for the object can then be described as follows:
\begin{align}
\frac{\partial U}{\partial \mathbf p^m} - \mathbf F &= \mathbf 0, \label{eq:equm} \\
\frac{\partial U}{\partial \mathbf p^f} &= \mathbf 0, \label{eq:equf} \\
\frac{\partial U}{\partial \mathbf p^u} &= \mathbf 0,
\end{align}
where $\mathbf F$ is the external force vector applied on the manipulated points. To solve the above equations, we need exact knowledge about that deformable object's deformation property, which is not available or difficult to acquire in many applications. To cope with this issue, we first simplify the potential energy function to only depend on $\mathbf p^m$ and $\mathbf p^f$, which is reasonable because usually the uninformed points are far from the manipulated and feedback points and thus their influence on the manipulation process is small and can be neglected. Next, we perform Taylor expansion of Equation~\ref{eq:equm} and Equation~\ref{eq:equf} about the current static equilibrium status $(\mathbf p^f_*, \mathbf p^m_*)$, and the equation of equilibrium implies a relationship between the relative displacements of feedback points and the manipulated points:
\begin{equation}
A(\delta \mathbf p^f) + B(\delta \mathbf p^m) = \mathbf 0,
\end{equation}
where $\delta \mathbf p^f = \mathbf p^f -\mathbf p^f_*$ and $\delta \mathbf p^m = \mathbf p^m-\mathbf p^m_*$ are the displacement relative to the equilibrium for feedback points and manipulated points, respectively. The functions $A(\cdot)$ and $B(\cdot)$ are nonlinear in general, though they can be linear in some special cases. For instance, when only performing the first order Taylor expansion as what is done in~\cite{Alarcon:2016:ADM}, $A(\delta \mathbf p^f) = \frac{\partial^2 U}{\partial \mathbf p^m \partial \mathbf p^f}$ and $B(\delta \mathbf p^m) =  \frac{\partial^2 U}{\partial (\mathbf p^m)^2} \delta \mathbf p^m$ are two linear functions. In this paper, we allow $A(\cdot)$ and $B(\cdot)$ to be general smooth linear functions to estimate a better model for the deformable object manipulation process. 

We further assume the function $B(\cdot)$ to be invertible, which implies
\begin{equation}
\label{eq:D}
\delta \mathbf p^m = D(\delta \mathbf p^f),
\end{equation}
where $D = A \circ B^{-1}$ is the mapping between the velocities of the feedback points and the manipulated points. In this way,  we can determine a suitable end-effector velocity via feedback control $\delta_p^m =  D(\eta \cdot \Delta \mathbf p^f)$ to derive the object toward its goal state, where $\Delta \mathbf p^f = \mathbf p^f_d - \mathbf p^f$ is the difference between the desired vector and the current vector of the feedback points and $\eta$ is the feedback gain. 

However, the velocities of feedback points usually cannot be directly used in the control, because in practice these velocities are measured using visual tracking of deformable objects and thus are likely to be noisy or even unreliable when tracking fails. More importantly, a soft object needs a large set of feedback points to characterize its deformation, but a robotic manipulation system usually only has a few end-effectors, and thus the $D(\cdot)$ function is Equation~\ref{eq:D} is a mapping from the high-dimensional space of feedback point velocities to the low-dimensional space of manipulated point velocities. Such a system is extremely underactuated and thus the convergence speed of the control would be slow. 

To deal with aforementioned difficulties, we extract a low-dimensional feature vector $\mathbf x$ from the feedback points for the control purpose, where $\mathbf x = C(\mathbf p^f)$ is a feature vector whose dimension is much smaller than that of $\mathbf p^f$, and the function $C(\cdot)$ is the feature extraction function. Around the equilibrium state, we have $\delta \mathbf x = C'(\mathbf p^f_*) \delta \mathbf p^f_*$, and can rewrite the equilibrium function using the feature vector as
\begin{equation}
\delta \mathbf p^m = D(C'(\mathbf p^f_*)^{-1} \delta \mathbf x) \triangleq H(\delta \mathbf x),
\end{equation}
where the function $H(\cdot)$ is called the deformation function. 

The manipulation problem of deformable object can finally be described as: given the desired state $\mathbf x_d$ of an object in the feature space, design a controller which learns the nonlinear function $H(\cdot)$ in an online manner, and outputs the control velocity $\delta \mathbf p^m$ decreasing the distance between the object's current state $\mathbf x$ and $\mathbf x_d$, i.e. $\delta \mathbf p^m =  H(\eta \cdot \Delta \mathbf x)$, where $\Delta \mathbf x = \mathbf x_d - \mathbf x$ and $\eta$ is the feedback gain.
\section{Controller Design}
\label{sec:controllerdesign}
\subsection{Deformation Function Learning}
\label{sec:ldf}
Since the deformation function $H(\cdot)$ is a general and highly nonlinear function determining how the movement of the manipulated points is converted into the feature space, the learning of the function $H$ requires a flexible and non-parametric method. Our solution is to use the Gaussian Process Regression (GPR) to fit the deformation function in an online manner.  

GPR is a nonparametric regression technique which defines a distribution over functions and the inference takes place directly in the functional space given the covariance and mean of the functional distribution. For our manipulation problem, we formulate the deformation function $H$ as a Gaussian process:
\begin{equation}
	H \sim GP \left( m(\delta \mathbf x),\,\, k(\delta \mathbf x, \delta \mathbf x') \right)
\end{equation}
where $\delta \mathbf x$ still denotes the velocity in the feature space.  For the covariance or kernel function $k(\delta \mathbf x,\delta \mathbf x')$, we are using the Radius Basis Function (RBF) kernel: $k(\delta \mathbf x,\delta \mathbf x') = \exp(-\frac{\|\delta \mathbf x- \delta \mathbf x'\|^2}{2\sigma_{RBF}^2})$, where the parameter $\sigma_{RBF}$ sets the spread of the kernel. For the mean function $m(\delta \mathbf x)$, we are using the linear mean function $m(\delta \mathbf x) = \mathbf W \delta \mathbf x$, $\mathbf W$ is the linear regression weight matrix. We choose to use a linear mean function rather than the common zero mean function, because previous work~\cite{Alarcon:2016:ADM} showed that a linear function can capture a large part of the deformation function $H$. As a result, a linear mean function can result in faster convergence of our online learning process and also provide a relatively accurate prediction in the unexplored region in the feature space. The matrix $W$ is learned online by minimizing a squared error $Q = \frac{1}{2}\|\delta \mathbf p^m-\mathbf W\delta \mathbf x\|_2$ with respect to the weight matrix $\mathbf W$. 

Given a set of training data in terms of pairs of feature space velocities and manipulated point velocities $\{(\delta \mathbf x_t, \delta \mathbf p^m_t)\}_{t=1}^N$ during the previous manipulation process, the standard GPR
computes the distribution of the deformation function as a Gaussian process $H(\delta \mathbf x) \sim \mathcal N(\mu(\delta \mathbf x), \sigma^2(\delta \mathbf x))$, where GP's mean function is 
\begin{align}
\label{eq:mu}
\mu(\delta \mathbf x) &= m(\delta \mathbf x) + \mathbf k^T(\delta \mathbf X, \delta \mathbf x)\cdot [\mathbf K(\delta \mathbf X, \delta \mathbf X) + \sigma_n^2 \mathbf I]^{-1}\notag \\
&\quad\cdot (\delta \mathbf P^m -m(\delta \mathbf X)) 
\end{align}
and GP's covariance function is 
\begin{align}
\label{eq:sigma}
\sigma^2(\delta \mathbf x) &= k(\delta \mathbf x,\delta \mathbf x) - \mathbf k^T(\delta \mathbf X,\delta \mathbf x)\cdot [\mathbf K(\delta \mathbf X,\delta \mathbf X) + \sigma_n^2 \mathbf I]^{-1}  \notag\\
&\quad \cdot \mathbf k(\delta \mathbf X, \delta \mathbf x). 
\end{align}
Here $\delta \mathbf X$ and $\delta \mathbf P^m$ are matrices corresponding to the stack of $\{\delta \mathbf x_t\}_{t=1}^N$ and $\{\delta \mathbf p^m_t\}_{t=1}^N$ in the training data, respectively. $\mathbf K$ and $\mathbf k$ are matrices and vectors computed using a given covariance function $k(\cdot, \cdot)$. The matrix $\mathbf A = \mathbf K + \sigma_n^2 \mathbf I$ is called the Gram matrix, and the parameter $\sigma_n$ estimates the uncertainty or noise level of the training data. 

  

\subsection{Real-time Online GPR}
\label{sec:FO-GPR}

In the deformation object manipulation process, the data $(\delta \mathbf x_t, \delta \mathbf p_t^m)$ is generated sequentially, and thus at each time step $t$, we need to update the GP deformation function $H_t(\delta \mathbf x) \sim \mathcal N(\mu_t(\delta \mathbf x), \sigma^2(\delta \mathbf x))$ at an interactive manner, with
\begin{align}
\label{eq:mut}
\mu_t(\delta \mathbf x) &= m(\delta \mathbf x) + \mathbf k^T(\delta \mathbf X_t, \delta \mathbf x) \cdot [\mathbf K(\delta \mathbf X_t, \delta \mathbf X_t) + \sigma_n^2 \mathbf I]^{-1} \notag \\
&\quad \cdot (\delta \mathbf P^m_t -m(\delta \mathbf X_t))
\end{align}
and
\begin{align}
\label{eq:sigmat}
\sigma^2_t(\delta \mathbf x) &= k(\delta \mathbf x,\delta \mathbf x) - \mathbf k^T(\delta \mathbf X_t,\delta \mathbf x)\cdot [\mathbf K(\delta \mathbf X_t,\delta \mathbf X_t) + \sigma_n^2 \mathbf I]^{-1} \notag \\
&\quad \cdot \mathbf k(\delta \mathbf X_t, \delta \mathbf x).
\end{align}

In the online GPR, we need to perform the inversion of the Gram matrix $\mathbf A_t = \mathbf K(\delta \mathbf X_t, \delta \mathbf X_t) + \sigma_n^2 \mathbf I$ repeatedly with a time complexity $\mathcal O(N^3)$, where $N$ is the size of the current training set involved in the regression. Such cubic complexity makes the training process slow for long manipulation sequence where the training data size $N$ increases quickly. In addition, the growing up of the GP model will reduce the newest data's impact on the regression result and make the GP fail to capture the change of the objects's deformation parameters during the manipulation. This is critical for deformable object manipulation, because the deformation function $H$ is derived from the local force equilibrium and thus is only accurate in a small region.


Motivated by previous work about efficient offline GPR~\cite{Snelson:2006:SGP, Rasmussen:2002:MGP, Snelson:2007:GPA, Duy:2009:LGP}, we here present a novel online GPR method called the \emph{Fast Online GPR} (FO-GPR) to reduce the high computational cost and to adapt to the changing deformation properties while updating the deformation model during the manipulation process. The main idea of FO-GPR includes two parts: 1) maintaining the inversion of the Gram matrix $\mathcal A_t$ incrementally rather using direct matrix inversion; 2) restricting the size of $\mathbf A_t$ to be smaller than a given size $M$, and $\mathbf A_t$'s size exceeds that limit, using a selective ``forgetting'' method to replace stale or uninformative data by the fresh new data point. 

\subsubsection{Incremental Update of Gram matrix $\mathbf A_t$} Suppose at time $t$, the size of $\mathbf A_t$ is still smaller than the limit $M$. In this case, $A_t$ and $A_{t-1}$ are related by  
\begin{equation}
	\mathbf A_t = 
  \begin{bmatrix}
	\mathbf A_{t-1} & \mathbf b \\ \mathbf b^T & c
  \end{bmatrix},
\end{equation}
where $\mathbf b = \mathbf k(\delta \mathbf X_{t-1},\delta \mathbf x_t)$ and $c = k(\delta \mathbf x_t,\delta \mathbf x_t) + \sigma^2_n$. According to the Helmert-–Wolf blocking inverse property, we can compute the inverse of $\mathbf A_t$ based on the inverse of  $\mathbf A_{t-1}$:
\begin{align}
\label{eq:incremA}
	\mathbf A_t^{-1} & = 
    \begin{bmatrix}
    \left(\mathbf A_{t-1}-\frac{1}{c}\mathbf {bb}^T \right)^{-1} & -\frac{1}{r}\mathbf A_{t-1}^{-1} \mathbf b \\ -\frac{1}{r}\mathbf b^T\mathbf A_{t-1}^{-1} & \frac{1}{r}
    \end{bmatrix} \notag \\
    & = 
    \begin{bmatrix}
    \mathbf A_{t-1}^{-1} + \frac{1}{r} \mathbf A_{t-1}^{-1} \mathbf {bb}^T\mathbf A_{t-1}^{-1} & -\frac{1}{r}\mathbf A_{t-1}^{-1} \mathbf b \\ -\frac{1}{r}\mathbf A_{t-1}^T\mathbf A_{t-1}^{-1} & \frac{1}{r}
    \end{bmatrix},
\end{align}
where $r = c-\mathbf b^T\mathbf A_{t-1}^{-1}\mathbf b$. In this way, we achieve the incremental update of the inverse Gram matrix from $\mathbf A_{t-1}^{-1}$ to $\mathbf A_t^{-1}$, and the computational cost is $\mathcal O(N^2)$ rather than $\mathcal O(N^3)$ of direct matrix inversion. This acceleration enables fast GP model update during the manipulation process.

\subsubsection{Selective Forgetting in Online GPR} When the size of $\mathbf A_{t-1}$ reaches the limit $M$, we use a ``forgetting'' strategy to replace the most uninformative data by the fresh data points while keeping the size of $\mathbf A_t$ to be $M$. In particular, we choose to forget the $i_*$ data point that is the most similar to other data points in terms of the covariance, i.e.,  
\begin{equation}
\label{eq:opti}
 i_* = \argmax_{i} \sum\nolimits_j\mathbf A[i, j],
\end{equation}
where $\mathbf A[i,j]$ denotes the covariance value stored in the $i$-th row and $j$-th column in $\mathbf A$, i.e., $k(\delta \mathbf x_i, \delta \mathbf x_j) + \sigma_n^2$.

Given the new data $(\delta \mathbf x_t, \delta \mathbf p^m_t)$, we need to update $\delta \mathbf X_t$, $\delta \mathbf P^m_t$, and $\mathbf A_t^{-1} = [\mathbf K(\delta \mathbf X_t, \delta \mathbf X_t) + \sigma_n^2 \mathbf I]^{-1}$ in Equation~\ref{eq:mut} and~\ref{eq:sigmat} by swapping data terms related to $\delta \mathbf x_t$ and $\delta \mathbf x_{i_*}$, in order to update the deformation function $H_t$. 

The incremental update for $\delta \mathbf X_t$ and $\delta \mathbf P_m^t$ is trivial: $\delta \mathbf X_t$ is identical to $\delta \mathbf X_{t-1}$ except $\delta \mathbf X_t [i_*]$ is $\delta \mathbf x_t$ rather than $\delta \mathbf x_{i_*}$; $\delta \mathbf P_t^m$ is identical to $\delta \mathbf P_{t-1}^m$ except $\delta \mathbf P_t^m [i_*]$ is $\delta \mathbf p_t^m$ rather than $\delta \mathbf p_{i_*}^m$.

We then discuss how to update $\mathbf A_t$ from $\mathbf A_{t-1}$. Since $\mathbf A_t - \mathbf A_{t-1}$ is only non-zero at the $i_*$-th column or the $i_*$-th row:
\begin{equation}
(\mathbf A_t - \mathbf A_{t-1})[i,j]=\begin{cases}
0, & i,j \neq i_* \\
k(\delta \mathbf x_i, \delta \mathbf x) - k(\delta \mathbf x_i, \delta \mathbf x_{i_*}), & j = i_* \\
k(\delta \mathbf x_j, \delta \mathbf x) - k(\delta \mathbf x_j, \delta \mathbf x_{i_*}), & i = i_*,
\end{cases} \notag
\end{equation}
this matrix can be written as the multiplication of two matrices $\mathbf U$ and $\mathbf V$, i.e. $\mathbf A_t - \mathbf A_{t-1} = \mathbf U \mathbf V^T$, where
\begin{equation}
\mathbf U = 
\begin{bmatrix}
\mathbf e_{i_*} & (\mathbf I -\frac{1}{2} \mathbf e_{i_*} \mathbf e_{i_*}^T) (\mathbf k_t - \mathbf k_{t-1})
\end{bmatrix} \notag
\end{equation}
and 
\begin{equation}
\mathbf V = 
\begin{bmatrix}
(\mathbf I -\frac{1}{2} \mathbf e_{i_*} \mathbf e_{i_*}^T) (\mathbf k_t - \mathbf k_{t-1}) & \mathbf e_{i_*}
\end{bmatrix}. \notag
\end{equation}
Here $\mathbf e_{i_*}$ is a vector that is all zero but one at the $i_*$-th item, $\mathbf k_t$ is the vector $\mathbf k(\delta \mathbf X_{t}, \delta \mathbf x_t)$ and $\mathbf k_{t-1}$ is the vector $\mathbf k(\delta \mathbf X_{t-1}, \delta \mathbf x_{i_*})$. Both $\mathbf U$ and $\mathbf V$ are size $M \times 2$ matrices. 

Then using Sherman-Morrison formula, there is
\begin{align}
\label{eq:incremA2}
\mathbf A_t^{-1} &= (\mathbf A_{t-1} + \mathbf U \mathbf V^T)^{-1}\\
&= \mathbf A_{t-1}^{-1} - \mathbf A_{t-1}^{-1}\mathbf U (\mathbf I + \mathbf V^T \mathbf A_{t-1}^{-1} \mathbf U)^{-1} \mathbf V^T \mathbf A_{t-1}^{-1},  \notag 
\end{align}
which provides the incremental update scheme for the Gram matrix $\mathbf A_t$. Since $\mathbf {I+V}^T\mathbf A_{t-1}^{-1} \mathbf U$ is a $2\times 2$ matrix, its inversion can be computed in $\mathcal O(1)$ time. Therefore, the incremental update computation is dominated by the matrix-vector multiplication and thus the time complexity is $\mathcal O(M^2)$ rather than $\mathcal O(M^3)$. 

A complete description for FO-GPR is as shown in Algorithm~\ref{algo:FOGPR}.

\begin{algorithm}
 \caption{FO-GPR}
 \begin{algorithmic}[1]
 \label{algo:FOGPR}
 \renewcommand{\algorithmicrequire}{\textbf{Input:}}
 \renewcommand{\algorithmicensure}{\textbf{Output:}}
 \REQUIRE $\delta \mathbf X_{t-1}$, $\delta \mathbf P^m_{t-1}$, $\mathbf A_{t-1}$, $\delta \mathbf x_t$, $\delta \mathbf p^m_t$
 \ENSURE $\delta \mathbf X_t$, $\delta \mathbf P^m_t$, $\mathbf A_t^{-1}$ 
 \IF{$dim(\mathbf A_{t-1}) < M$}
 \STATE $\delta \mathbf X_t = [\delta \mathbf X_{t-1}, \delta \mathbf x_t]$ \\
 \STATE $\delta \mathbf P^m_t = [\delta \mathbf P^m_{t-1}, \delta \mathbf p_t^m]$ \\
 \STATE $\mathbf A_t^{-1}$ computed using Equation~\ref{eq:incremA}
 \ELSE
 \STATE $i_*$ computed using Equation~\ref{eq:opti} \\
 \STATE $\delta \mathbf X_t = \delta \mathbf X_{t-1}$, $\delta \mathbf X_t[i_*] = \delta \mathbf x_t$ \\
  \STATE $\delta \mathbf P_t^m = \delta \mathbf P_{t-1}^m$, $\delta \mathbf P_t^m[i_*] = \delta \mathbf p^m_t$ \\
  \STATE $\mathbf A_t^{-1}$ computed using Equation~\ref{eq:incremA2}
 \ENDIF
 \end{algorithmic} 
 \end{algorithm}
    
\subsection{Exploitation and Exploration}
\label{sec:exploration}
Given the deformation function $H_t$ learned by FO-GPR, the controller system predicts the required velocity to be executed by the end-effectors based on the error between the current state and the goal state in the feature space:
\begin{align}
\delta \mathbf p^m = H_t(\eta \cdot (\mathbf x^d - \mathbf x)),
\end{align}
where $\eta$ is a scale factor as the feedback gain. 

However, when there is no sufficient data, GPR cannot output control policy with high confidence, which typically happens in the early step of the manipulation or when the robot manipulates the object into a new unexplored configuration. Fortunately, the GPR framework provides a natural way to trade-off exploitation and exploration by sampling the control velocity from distribution of $H_t$:
\begin{equation}
	\delta \mathbf p^m \sim \mathcal N(\mu_t, \sigma_t^2)
\end{equation}
If $\mathbf x_t$ is now in unexplored region with large $\sigma_t^2$, the controller will perform exploration around $\mu_t$; if $\mathbf x_t$ is in a well-explored region with small $\sigma_t^2$, the controller will output velocity close to $\mu_t$. 

\begin{figure}[t] 
\centering
\includegraphics[width=1.0\linewidth]{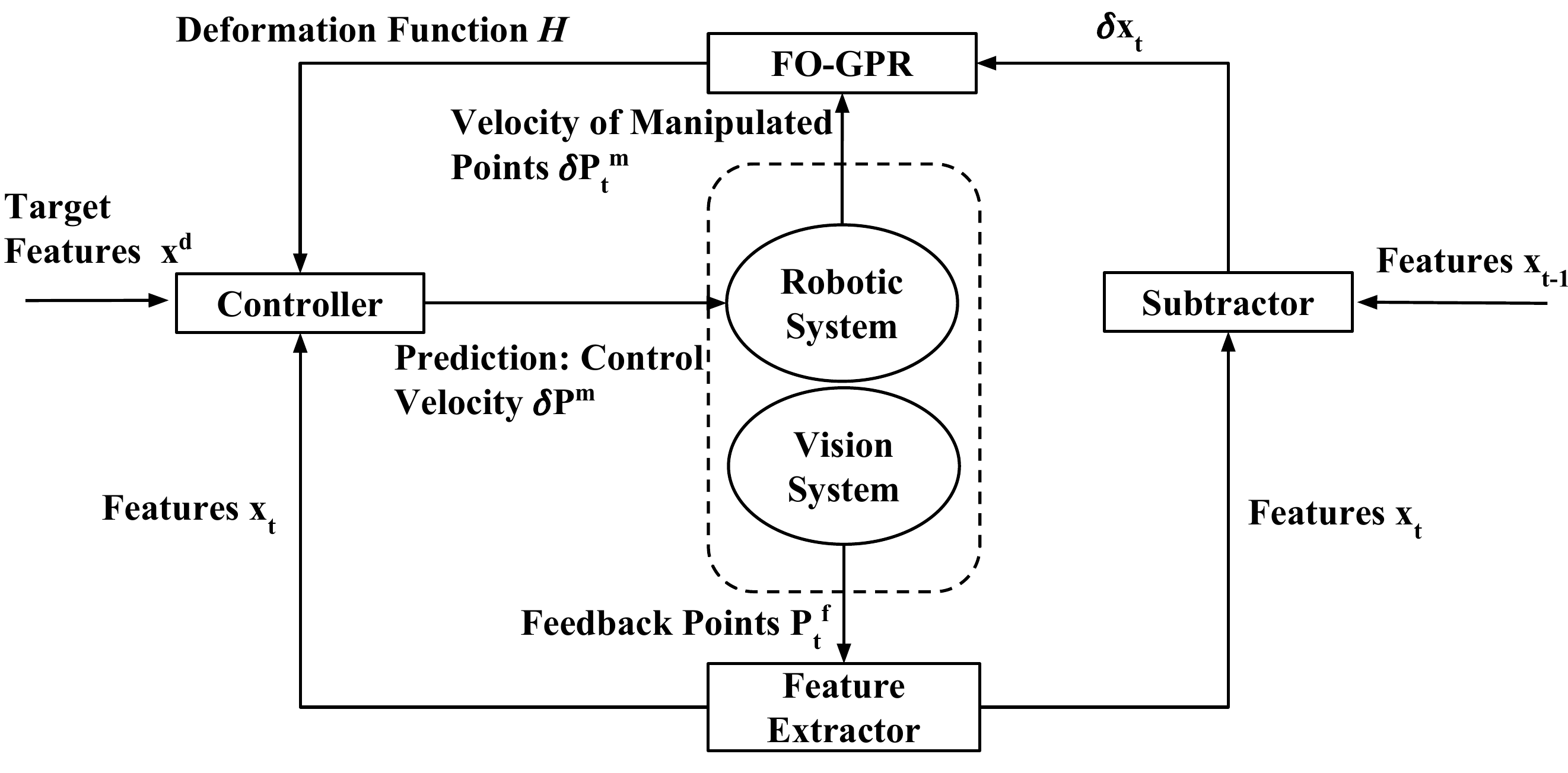}
\caption{An overview of our deformable object manipulation system. }
\label{fig:flow-chart}
\vspace*{-0.2in}
\end{figure}

A complete description of the controller based on FO-GPR is shown in Figure~\ref{fig:flow-chart}.
 
\section{Feature Extraction}
\label{sec:featureextraction}
For rigid body manipulation, an object's state can be completely described by its centroid position and orientation. But such global features are not sufficient to determine the configuration of a deformable object. As mentioned in Section~\ref{sec:controllerdesign}, we extract a feature vector $\mathbf x$ from the feedback points to represent the object's configuration. $\mathbf x$ is constituted of two parts in terms of global and local features. 

\subsection{Global Features}
\label{sec:globalfeatures}

\subsubsection{Centroid} The centroid feature $\mathbf c \in \mathcal R^3$ is computed as the geometric center of the 3D coordinates of the feedback points：
\begin{equation}
\label{eq:centroid}
	\mathbf c = \left ( \mathbf p^f_1 + \mathbf p^f_2 + \cdots + \mathbf p^f_K \right ) / K
\end{equation}
and use $\mathbf c$ as part of $\mathbf x$ for the position term. When the number of feedback points $K$ increases, the estimation of the centroid is more accurate. The centroid feature is preferred when there are many feedback points.

\subsubsection{Positions of feedback points} Another way to describe a deformable object's configuration is to directly use the positions of all feedback points as part of $\mathbf x$, i.e. 
\begin{equation}
\label{eq:position}
	\mathbf \rho = \left[ (\mathbf p^f_1)^T, (\mathbf p^f_2)^T, \cdots, (\mathbf p^f_K)^T \right]
\end{equation}
This feature descriptor is advantageous when we want to drive all the feedback points toward their desired positions, but comes with a defect that its dimension increases rapidly when the number of feedback points to be tracked increases.


\subsection{Local Features}
\label{sec:localfeatures}

\subsubsection{Distance between points} The distance between each pair of feedback points intuitively measures the stretch of deformable objects. This feature is computed as 
\begin{equation}
\label{eq:distance}
	d = \| \mathbf p^f_1 - \mathbf p^f_2\|_2,
\end{equation}
where $\mathbf p^f_1$ and $\mathbf p^f_2$ are a pair of feedback points.

\subsubsection{Surface variation indicator} For deformable objects with developable surfaces, the surface variation around each feedback point can measure the local geometric property. Given a feedback point $\mathbf p$, we first compute the covariance matrix $\Omega$ for its neighborhood as 
\begin{equation}
	\Omega = \frac{1}{n}\sum_{i=1}^n\left(\mathbf p_i-\mathbf{\bar p}\right) \cdot \left(\mathbf p_i-\mathbf{\bar p}\right)^T，
\end{equation}
where $\{\mathbf p_i\}$ $(1\leq i\leq n)$ are surface points around $\mathbf p$ and $\mathbf{\bar p}=\frac{1}{n}\sum_{i=1}^n \mathbf p_i$ is the centroid of the neighboring points. The surface variation $\sigma$ is computed as
\begin{equation}
\label{eq:variation}
	\sigma = \frac{\lambda_0}{\lambda_0+\lambda_1+\lambda_2}
\end{equation}
where $\lambda_0$, $\lambda_1$, $\lambda_2$ are eigenvectors of $\Omega$ with $\lambda_0\leq\lambda_1\leq\lambda_2$. The variation indicator $\sigma$ needs sufficient surface sample points for accuracy.

\subsubsection{Extended FPFH from VFH} Extended FPFH is the local descriptor of VFH and is based on Fast Point Feature Histograms (FPFH)~\cite{Rusu:2009:FPF}. Its idea is to use a histogram to record differences between the centroid point $\mathbf p_c$ and its normal $\mathbf n_c$ with all other points and normals. Given a point $\mathbf p_i$ and its normal $\mathbf n_i$, we compute a Darboux coordinate frame with basis
\begin{equation}
	\begin{split}
	& \mathbf u_i = \mathbf n_c \\
    & \mathbf v_i = \frac{\mathbf p_i-\mathbf p_c}{\|\mathbf p_i-\mathbf p_c\|_2}\times \mathbf u_i \\
    & \mathbf w_i = \mathbf u_i \times \mathbf v_i
	\end{split}
\end{equation}
The differences between $\left\lbrace\mathbf p_c,\mathbf n_c \right\rbrace$ and $\left\lbrace\mathbf p_i,\mathbf n_i \right\rbrace$ are described by three values:
\begin{equation}
\label{eq:VFH_FPFH}
	\begin{split}
	& \cos(\alpha_i) = \mathbf v_i \cdot \mathbf n_i \\
    & \cos(\varphi_i) = \mathbf u_i \cdot \frac{\mathbf p_i - \mathbf p_c}{\| \mathbf p_i - \mathbf p_c \|_2} \\
    & \theta_i = \mathrm{atan2}(\mathbf w_i \cdot \mathbf n_i, \mathbf u_i \cdot \mathbf n_i)
	\end{split}
\end{equation}
These values are invariant to rotation and translation, making extended FPFH a useful local descriptor.

\section{Experiments and Results}
\label{sec:experiment}

\subsection{Experiment Setup}
\begin{figure}[t] 
\centering
\includegraphics[width=0.7\linewidth]{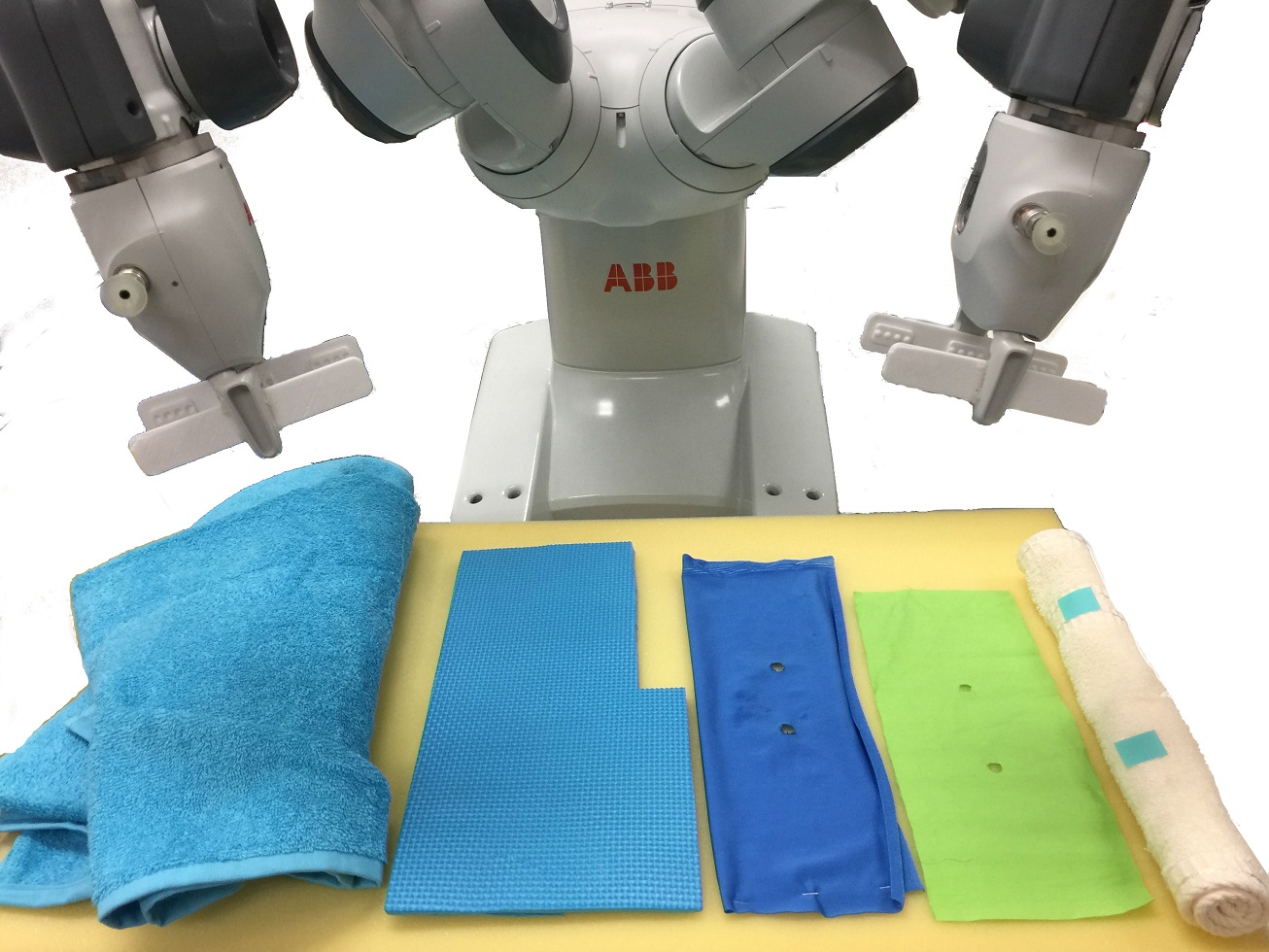}
\caption{The set of deformable objects to be manipulated in our experiment. From left to right: towel, plastic sheet, stretchable fabric, unstretchable fabric, and rolled towel.}
\label{fig:soft-objects}
\vspace*{-0.2in}
\end{figure}

We evaluate our approach using one dual-arm robot (ABB Yumi) with 7 degrees-of-freedom in each arm. To get precise information of the 3D object to be manipulated, we set up a vision system including two 3D cameras with different perception fields and precision: one realsense SR300 camera for small objects and one ZR300 camera for large objects. The entire manipulation system is shown in Figuree~\ref{fig:manipulation-system}. For FO-GPR parameters, we set the observation noise $\sigma_n = 0.001$, the RBF spread width $\sigma_{RBF} = 0.6$, and the maximum size of the Gram matrix $M = 300$. The executing rate of our approach is $30$ FPS. 

\begin{figure*}[]
\centering
\begin{subfigure}{0.18\textwidth}
\includegraphics[width=1.0\linewidth]{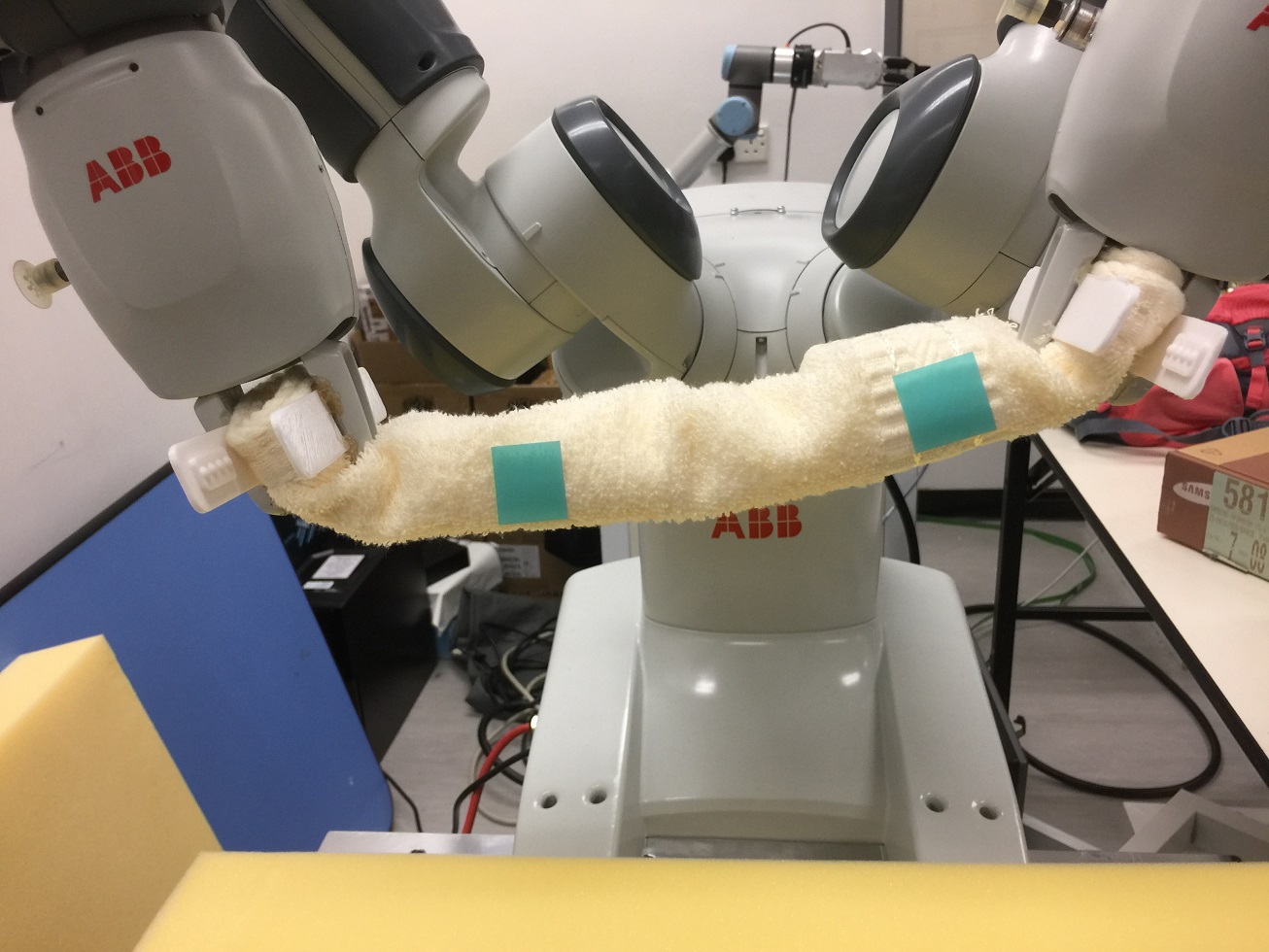}
\hspace{1in}
\end{subfigure}
\begin{subfigure}{0.18\textwidth}
\includegraphics[width=1.0\linewidth]{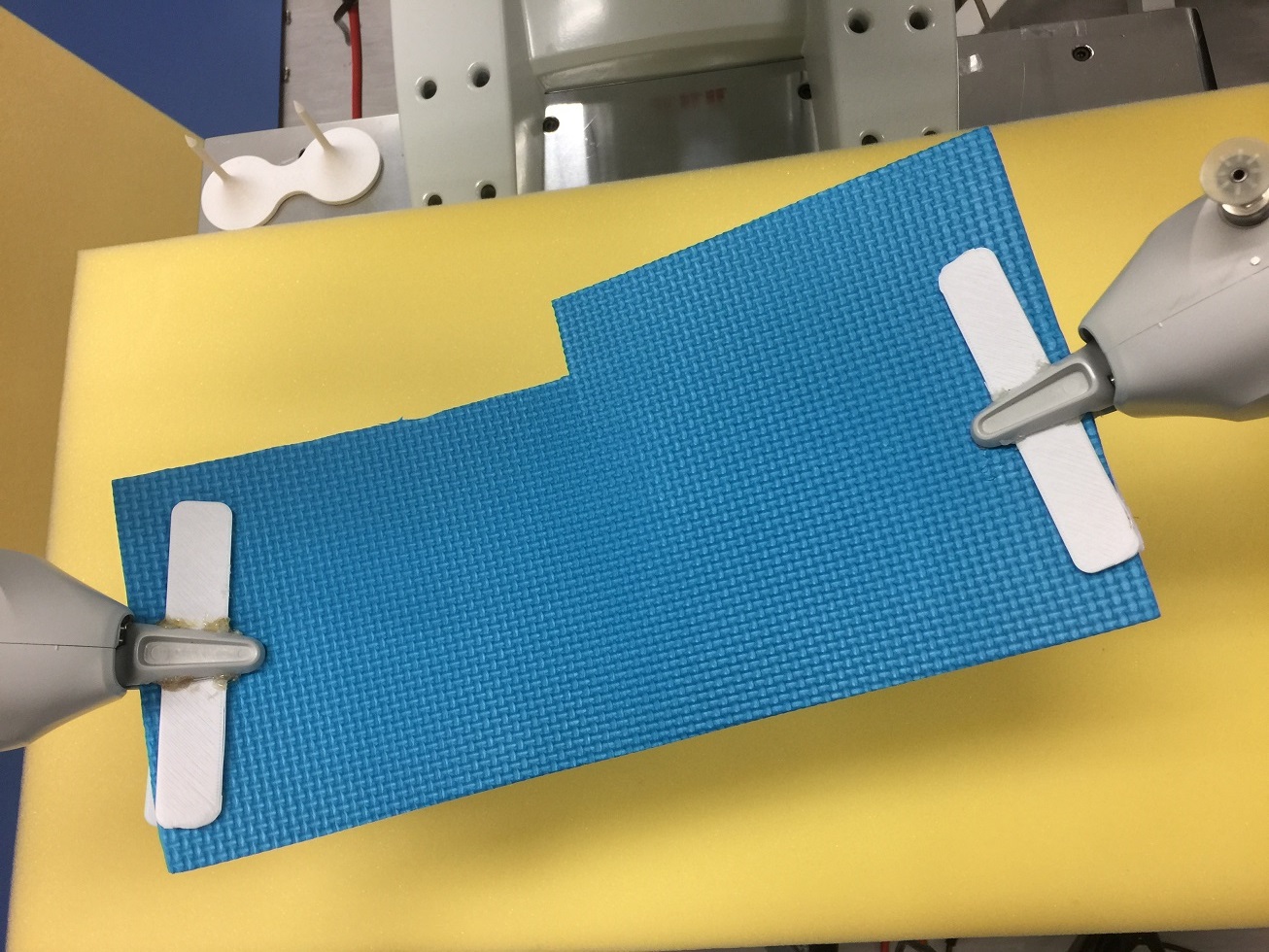}
\hspace{1in}
\end{subfigure}
\begin{subfigure}{0.18\textwidth}
\includegraphics[width=1.0\linewidth]{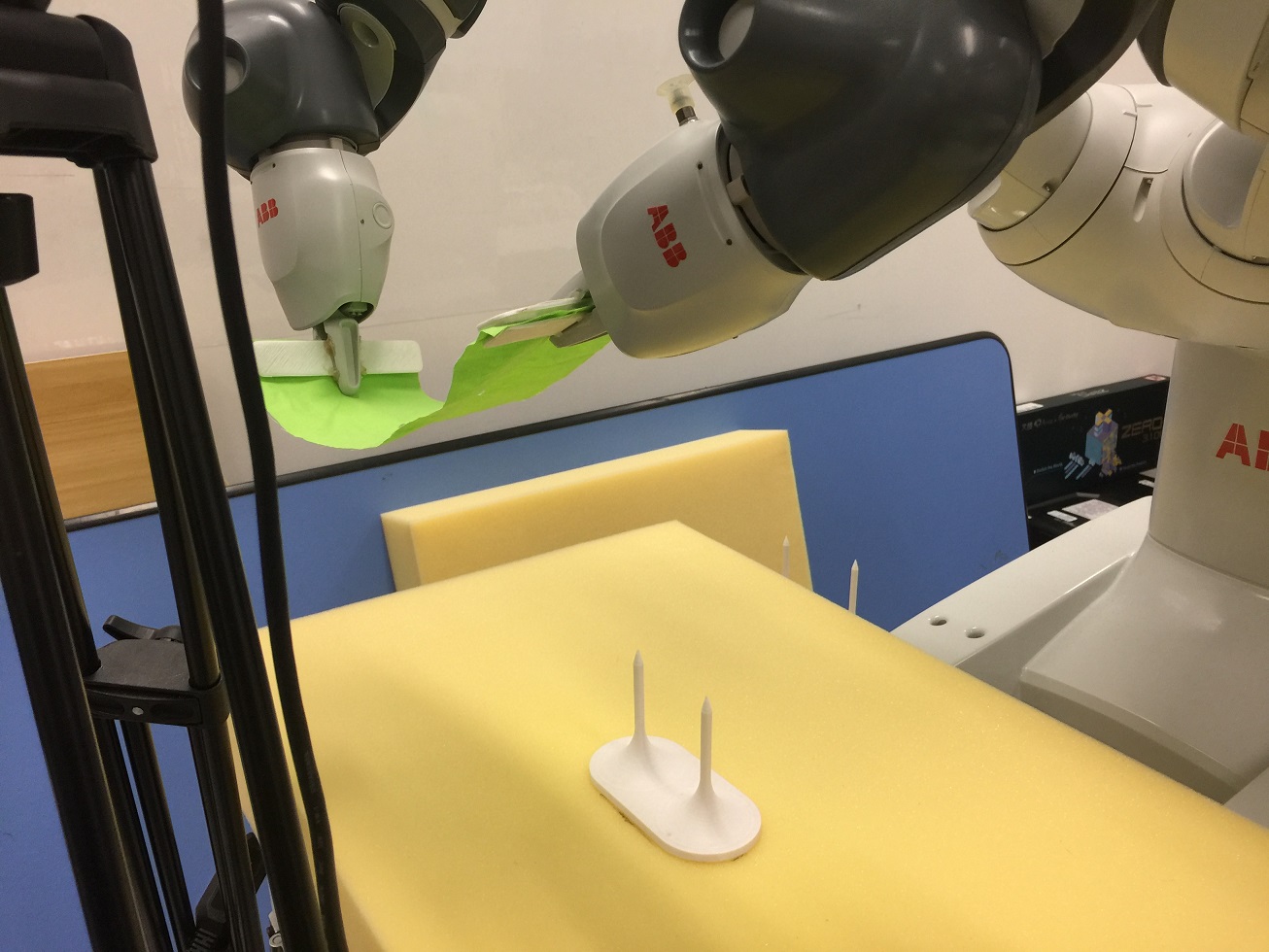}
\hspace{1in}
\end{subfigure}
\begin{subfigure}{0.18\textwidth}
\includegraphics[width=1.0\linewidth]{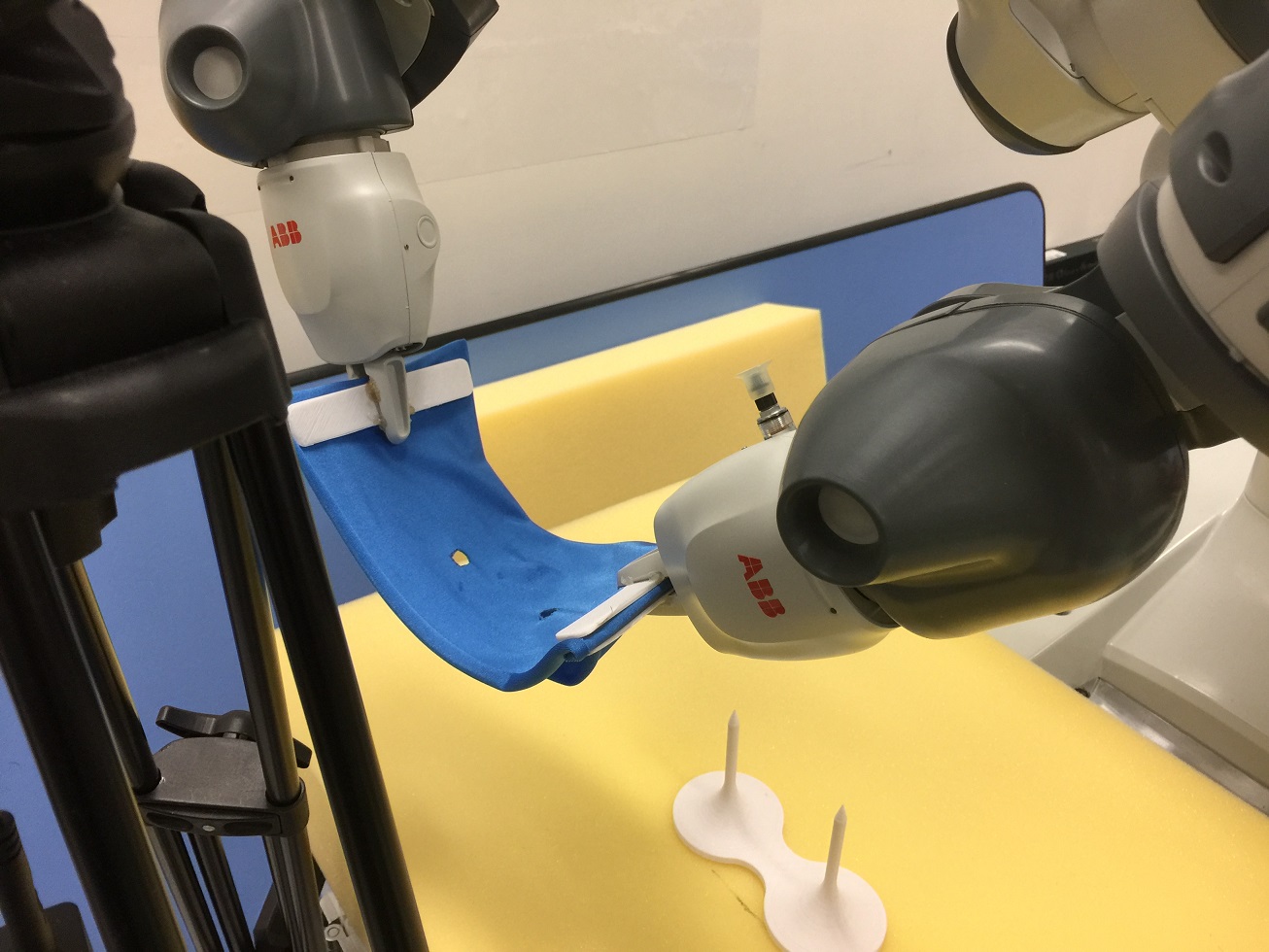}
\hspace{1in}
\end{subfigure}
\begin{subfigure}{0.18\textwidth}
\includegraphics[width=1.0\linewidth]{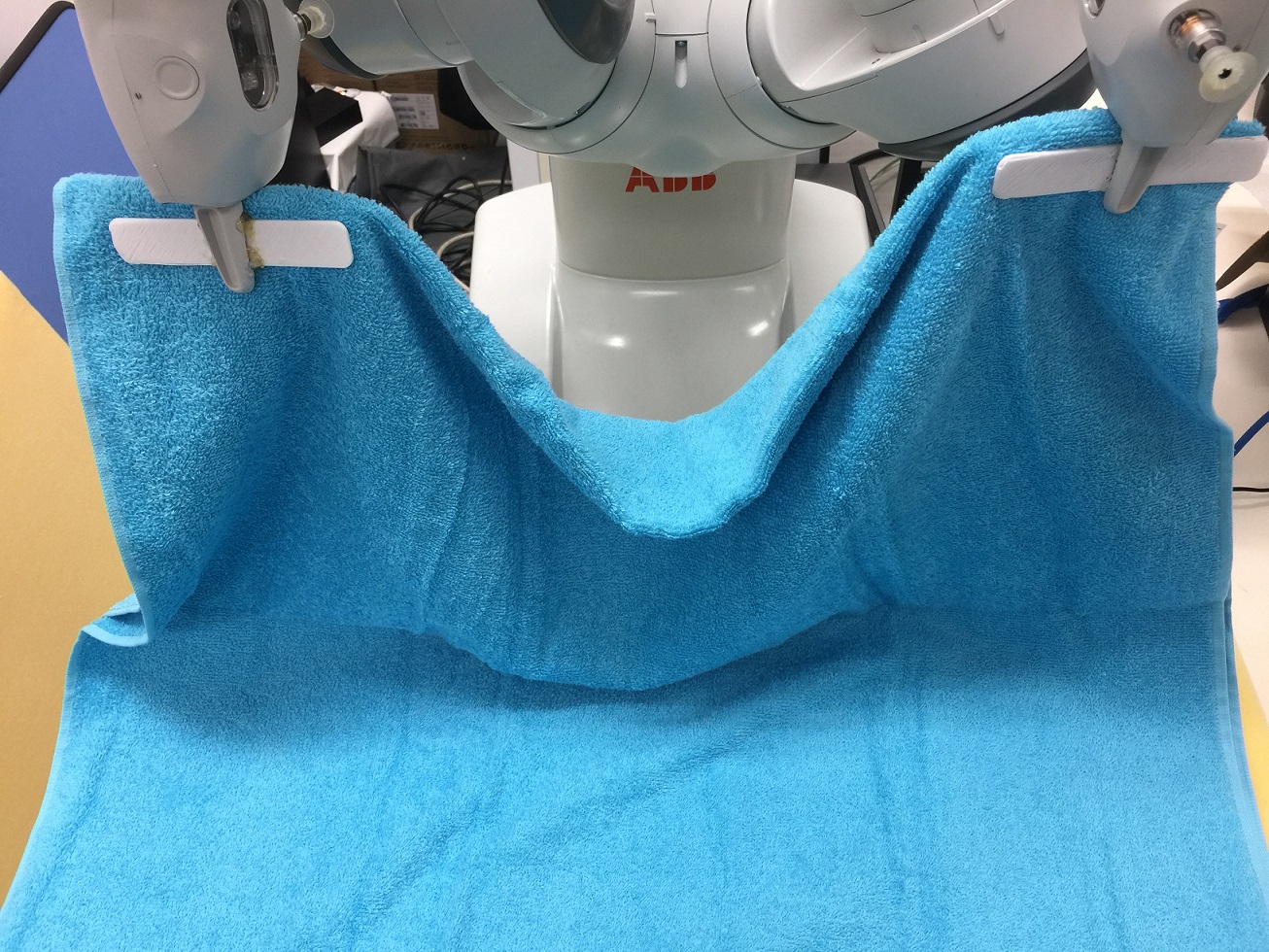}
\hspace{1in}
\end{subfigure}
\begin{subfigure}{0.18\textwidth}
\includegraphics[width=1.0\linewidth]{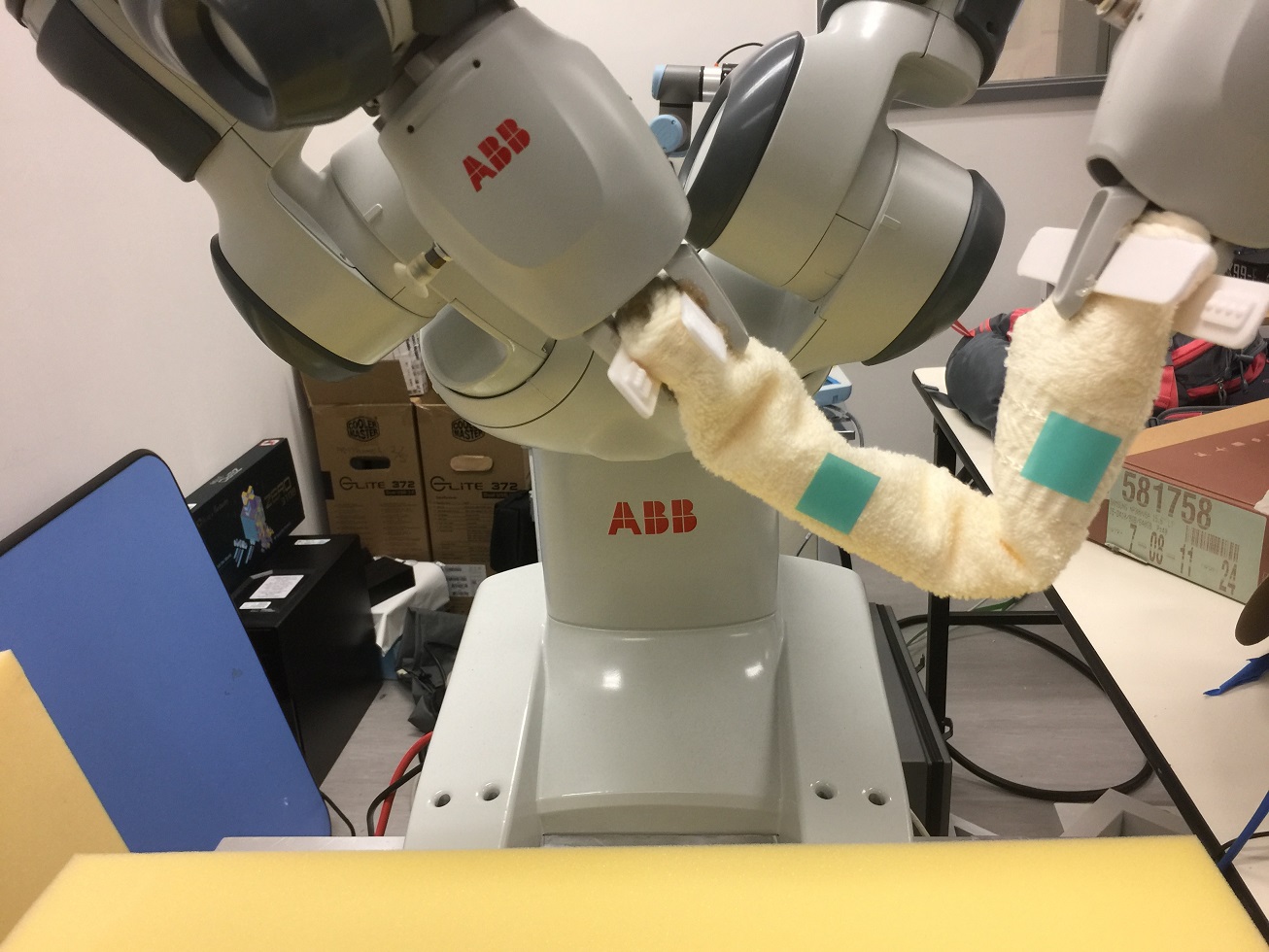}
\caption{}
\label{fig:rolled_towel}
\end{subfigure}
\begin{subfigure}{0.18\textwidth}
\includegraphics[width=1.0\linewidth]{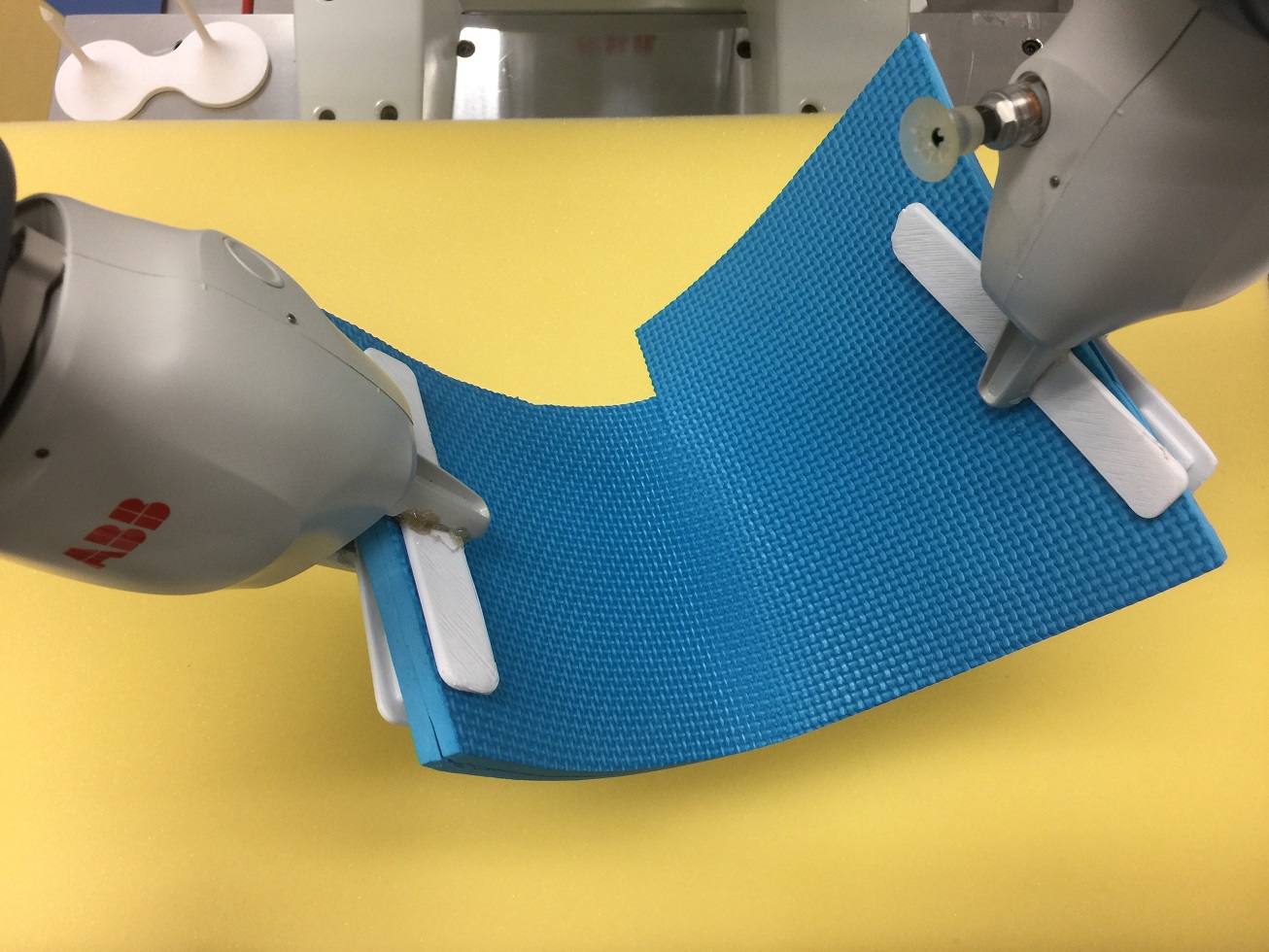}
\caption{}
\label{fig:plastic_plate}
\end{subfigure}
\begin{subfigure}{0.18\textwidth}
\includegraphics[width=1.0\linewidth]{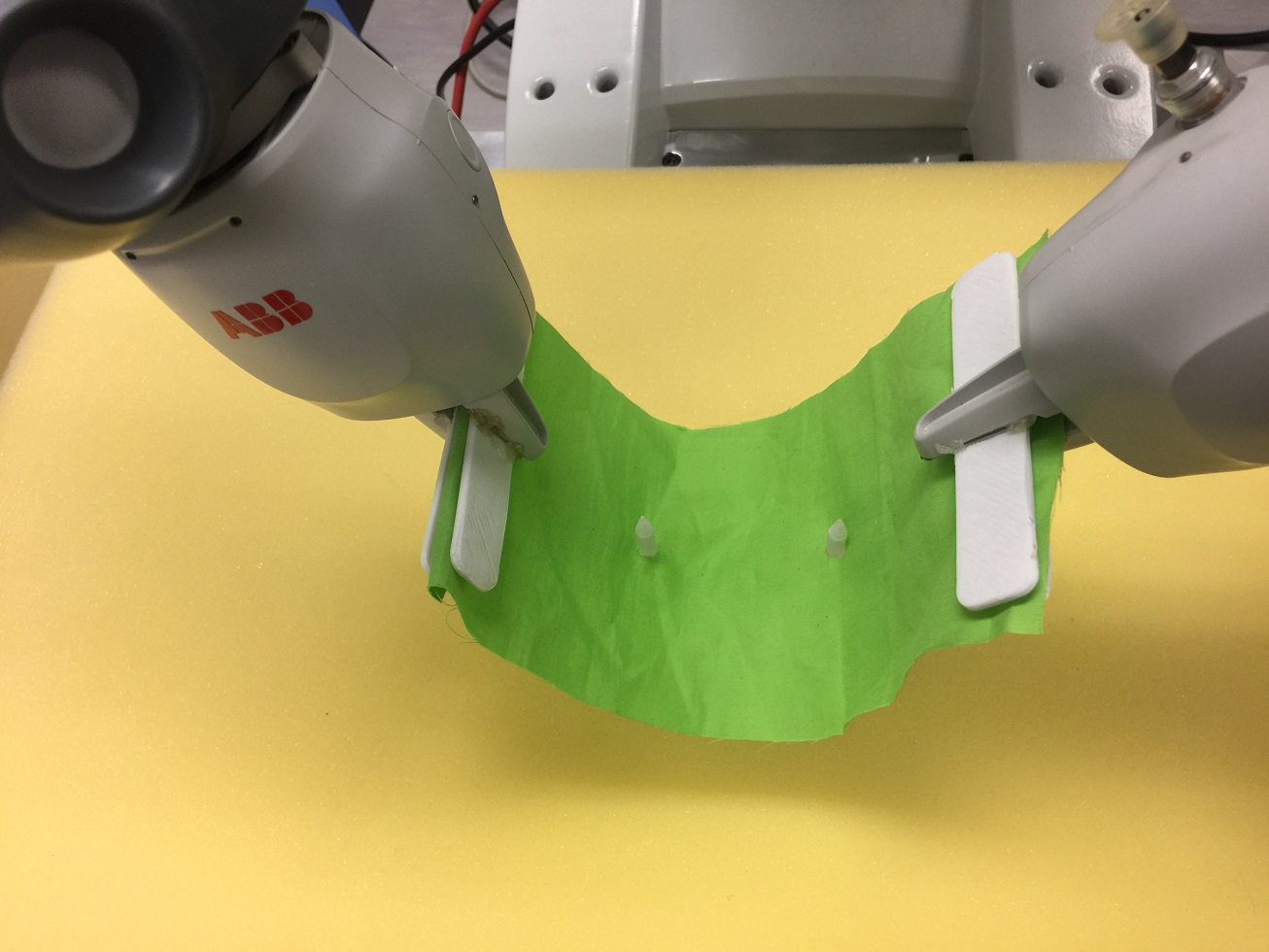}
\caption{}
\label{fig:stiff_fabric}
\end{subfigure}
\begin{subfigure}{0.18\textwidth}
\includegraphics[width=1.0\linewidth]{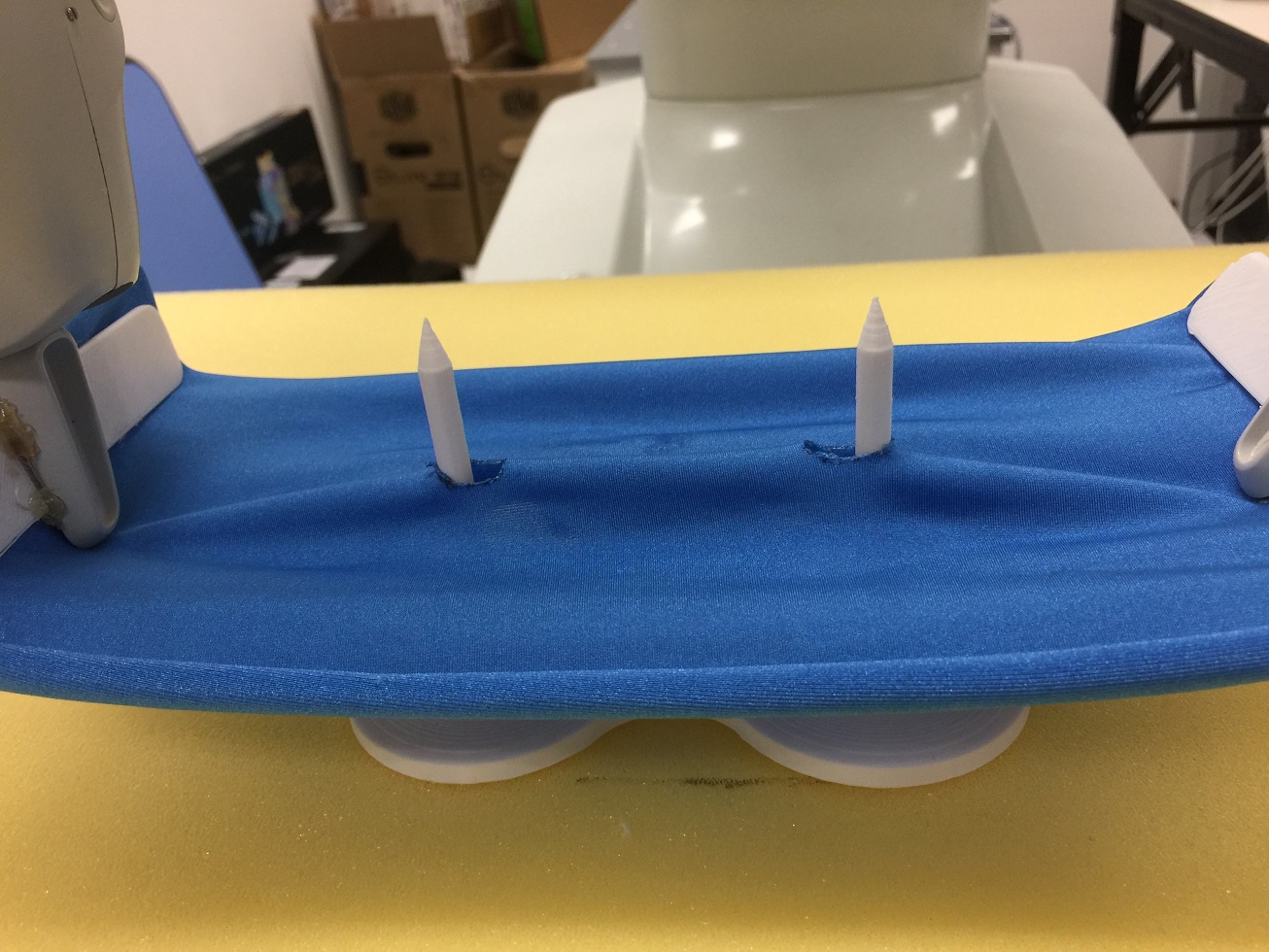}
\caption{}
\label{fig:elastic_fabric}
\end{subfigure}
\begin{subfigure}{0.18\textwidth}
\includegraphics[width=1.0\linewidth]{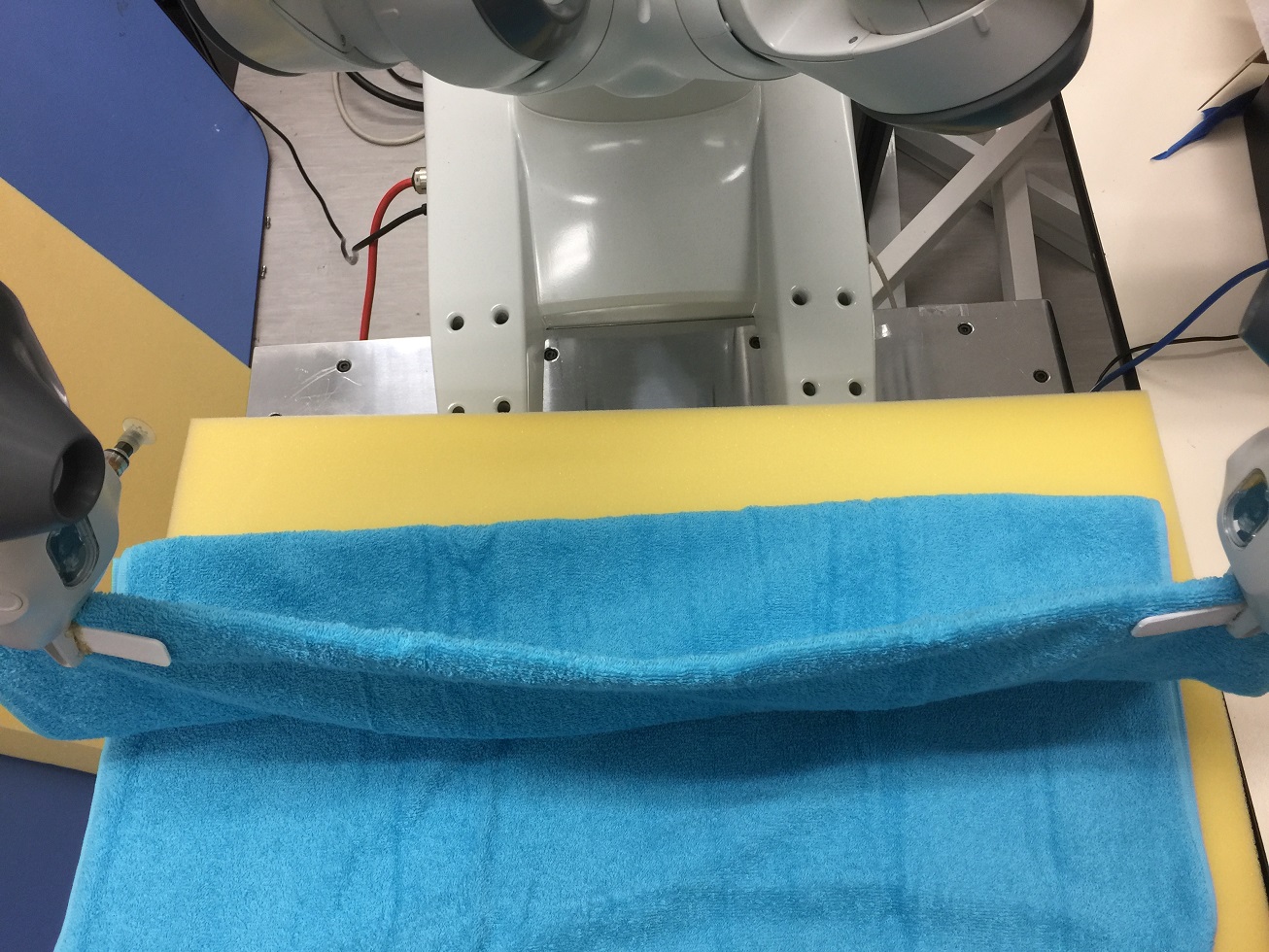}
\caption{}
\label{fig:bath_towel}
\end{subfigure}
\caption{The set of tasks used to evaluate the performance of our approach: (a) rolled towel bending, (b) plastic sheet bending, (c) peg-in-hole for unstretchable fabric, (d) peg-in-hole for stretchable fabric, and (e) towel folding. The first row shows the initial state of each object before the manipulation and the second row shows the goal states of the object after the successful manipulation.}
\label{fig:multi_tasks}
\vspace*{-0.3in}
\end{figure*}

\subsection{Manipulation Tasks}
To evaluate the performance of our approach, we apply it to different manipulation tasks involving distinct objects as shown in Figure~\ref{fig:soft-objects}. For each task, we choose different set of features as discussed in Section~\ref{sec:featureextraction} to achieve the best performance. 
\subsubsection{Rolled towel bending} This task aims at bending a rolled towel in to a specific goal shape as shown in Figure~\ref{fig:rolled_towel}. We use a $4$-dimension feature vector $\mathbf x=\left[\mathbf c, d\right]$ as the feature vector used in the FO-GPR driven visual-servo, where $\mathbf c$ is the centroid feature described by Equation~\ref{eq:centroid} and $d$ is the distance feature as described by Equation~\ref{eq:distance} for two feature points on the towel. 

\subsubsection{Plastic sheet bending}  This goal of this task is to manipulate a plastic sheet into a preassigned curved status as shown in Figure~\ref{fig:plastic_plate}. We use a $4$-dimension feature vector $\mathbf x=\left[\mathbf c, \sigma\right]$ to describe the state of the plastic sheet,  where $\mathbf c$ is the centroid feature described by Equation~\ref{eq:centroid} and $\sigma$ is the surface variation feature computed by Equation~\ref{eq:variation}.

\subsubsection{Peg-in-hole for fabrics} This task aims at moving cloth pieces so that the pins can be inserted into the corresponding holes on the fabric. Two different types of fabric with different stiffness have been tested in our experiment: one is an unstretchable fabric as shown in Figure~\ref{fig:stiff_fabric} and the other is a stretchable fabric as shown in Figure~\ref{fig:elastic_fabric}. The $6$-dimension feature vector $\mathbf x = \mathbf \rho$ is the position of feedback points as described in Equation~\ref{eq:position}. 

\subsubsection{Towel folding} This task aims at flattening and folding a towel into a desired state as shown in Figure~\ref{fig:bath_towel}. We use a binned histogram of extended FPFH to describe the towel's shape. The three values $\cos(\alpha)$, $\cos(\varphi)$ and $\theta$ are computed for all the feedback points using Equation~\ref{eq:VFH_FPFH}, and are then aggregated into $45$ bins individually, generating a feature vector of $135$ dimensions. Since the feature has a very large dimension, for this experiment we need to manually move the robot in the beginning to explore sufficient data so that the FO-GPR can learn a good enough initial model for the complex deformation function. 

\begin{figure}[t] 
\centering
\begin{subfigure}{0.22\textwidth}
\includegraphics[width=1.0\linewidth]{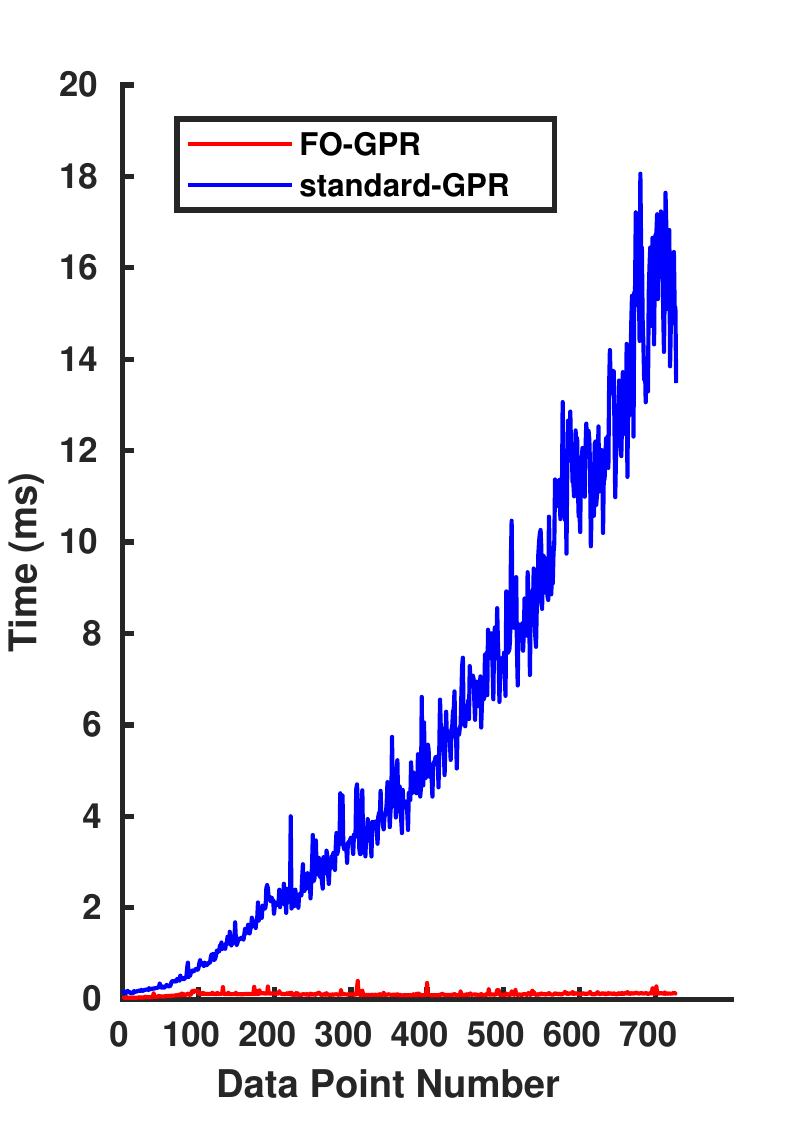}
\caption{}
\label{fig:idle-time}
\end{subfigure}
\begin{subfigure}{0.22\textwidth}
\includegraphics[width=1.0\linewidth]{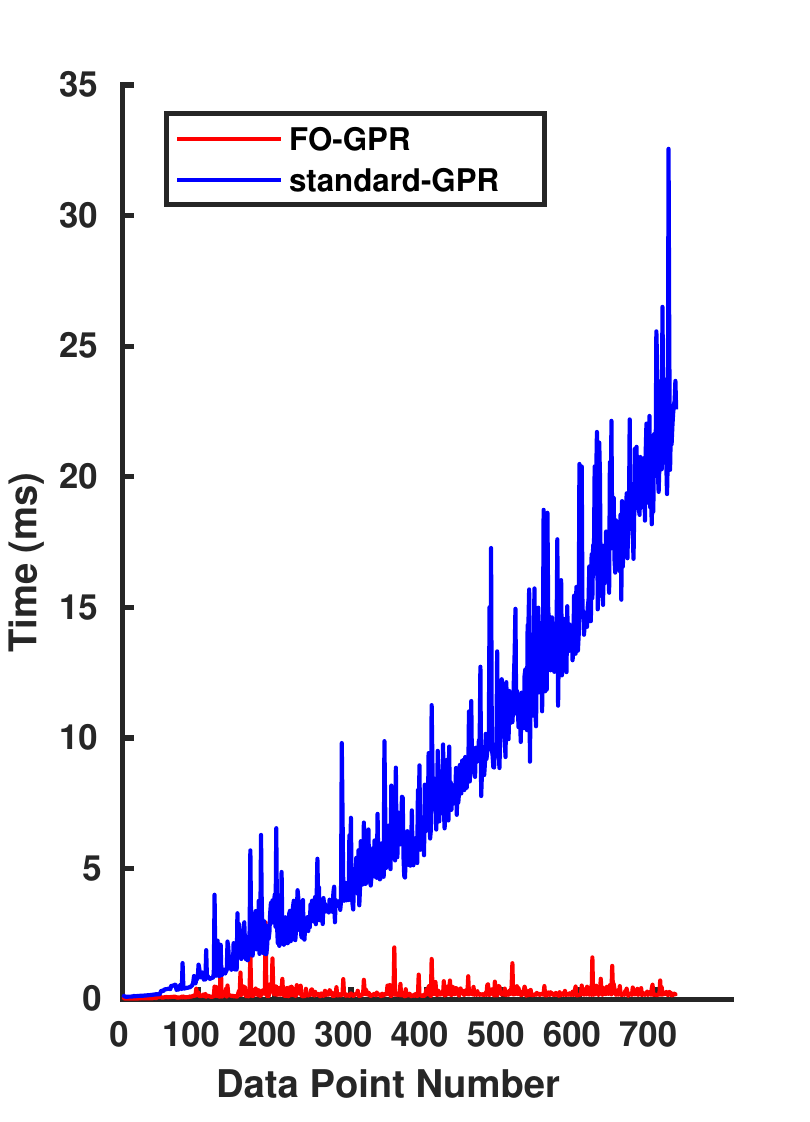}
\caption{}
\label{fig:busy-time}
\end{subfigure}
\caption{Comparison of the time cost of FO-GPR and standard GPR: (a) the time cost comparison between GP model estimation; (b) the time cost comparison for the entire manipulation process.}
\label{fig:time-comparison}
\vspace*{-0.2in}
\end{figure}

\begin{figure}[t] 
\centering
\includegraphics[width=1.0\linewidth]{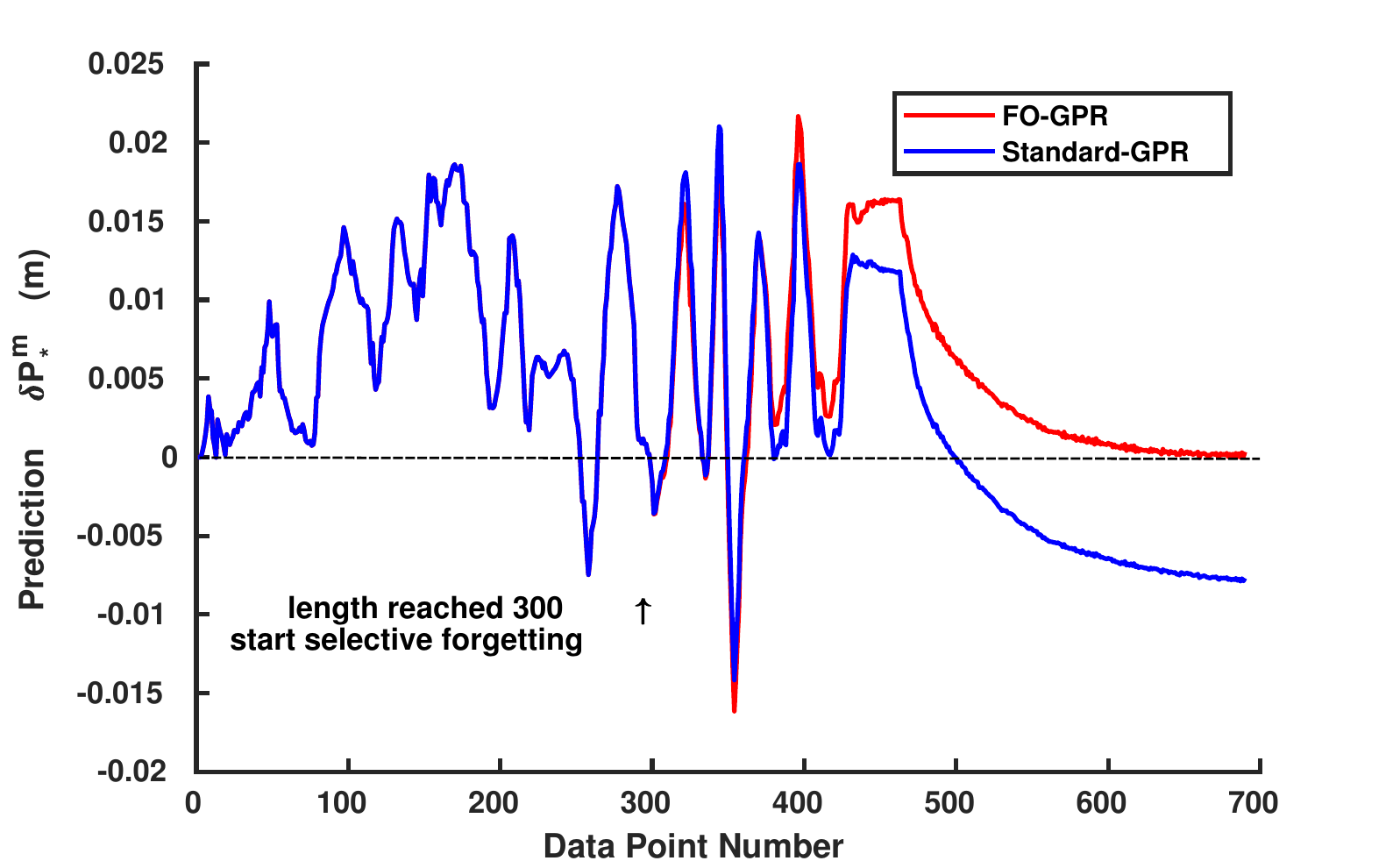}
\caption{Impact of selective forgetting in FO-GPR: FO-GPR is superior over the standard GPR in terms of the computational cost and the deformation model accuracy.
}
\label{fig:information}
\vspace*{-0.2in}
\end{figure}

\begin{figure}[t] 
\centering
\includegraphics[width=1.0\linewidth]{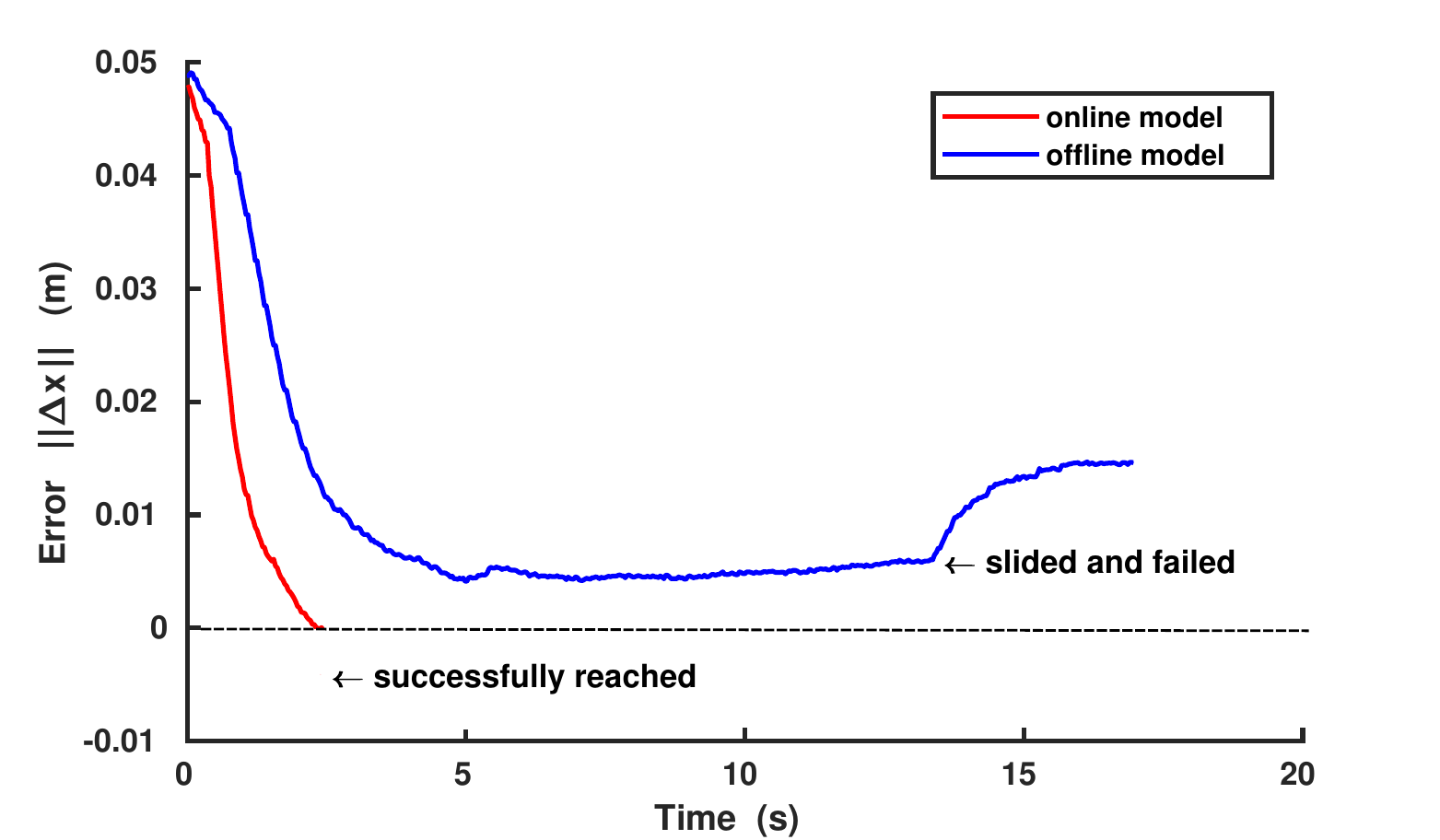}
\caption{Comparison of online and offline GP models for rolled towel manipulation. The controller based on the online model succeeds while the controller based on the offline model fails.}
\label{fig:online_offline}
\vspace*{-0.2in}
\end{figure}

\begin{figure}[t] 
\centering
\includegraphics[width=1.0\linewidth]{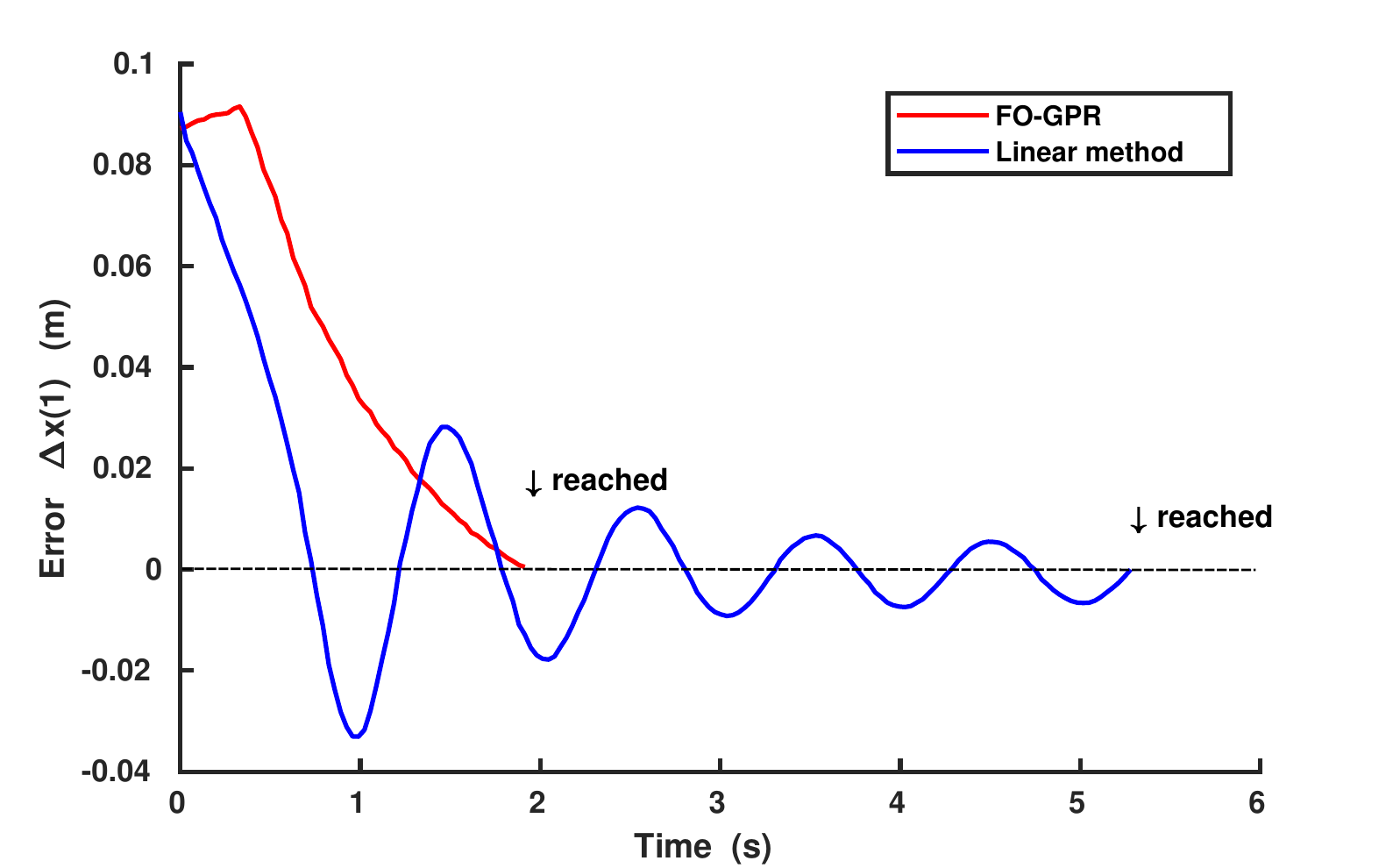}
\caption{Comparison of the controllers based on FO-GPR and the linear model on the rolled towel task.
}
\label{fig:FO_Linear4d1}
\vspace*{-0.2in}
\end{figure}

\begin{figure}[t] 
\centering
\includegraphics[width=1.0\linewidth]{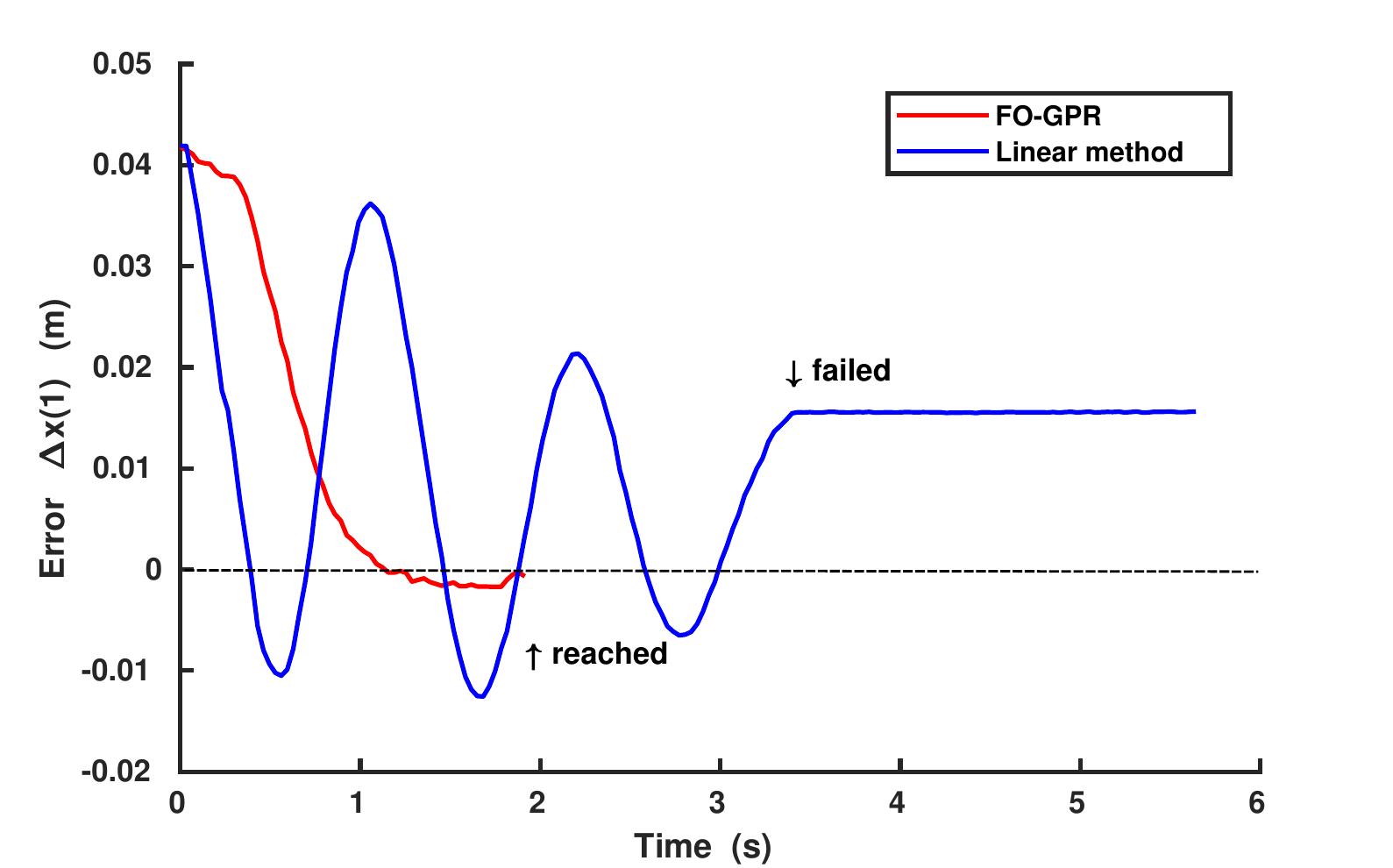}
\caption{Comparison of the controllers based on FO-GPR and the linear model on the peg-in-hole task with stretchable fabric.
}
\label{fig:FO_Linear6d}
\vspace*{-0.2in}
\end{figure}

\subsection{Results and Discussion}
\urlstyle{rm}
Our FO-GPR based manipulation control is able to accomplish all the five tasks efficiently and accurately. Please refer to the videos at \url{ https://sites.google.com/view/mso-fogpr} for manipulation details. 

Next, we provide some quantitative analysis of our approach by comparing with some state-of-the-art approaches. 

\subsubsection{Comparison of computational cost with standard GPR}
As shown in Figure~\ref{fig:idle-time}, the time cost of the standard GPR operation in each iteration increases significantly when the number of training points increases, which makes the online deformation function update impossible. Our FO-GPR method's time cost is always under $2$ ms. We also compare the time cost of each complete cycle of the manipulation process, including feature extraction, tracking, robot control, and GPR, and the result is shown in Figure~\ref{fig:busy-time}. Again, the time cost of manipulation using standard GPR fluctuates significantly, which can be 10 times slower than our FO-GPR based manipulation, whose time cost is always below $5$ ms and allows for real-time manipulation. This experiment is performed using the rolled towel bending task.

\subsubsection{Impact of selective forgetting}
In Figure~\ref{fig:information}, we compares the GP prediction quality between FO-GPR and the standard GPR on the rolled towel task, in order to show the impact of selective forgetting in FO-GPR. We record about $700$ data entries continuously. The first $450$ data are produced using random controller, and the rest are generated by the FO-GPR based controller which drives the soft object toward the target state smoothly. Before the data size reaches the maximum limit $M = 300$, the controllers using two GPR models provide the same velocity output. From this point on, FO-GPR selectively forgets uninformative data while the standard GPR still uses all data for prediction. For data points with indices between $300$ and $450$, the output from two controllers are similar , which implies that FO-GPR still provides a sufficiently accurate model. After that, the FO-GPR based controller drives the object toward goal and eventually the controller output is zero; while for standard GPR, the controller output remains unzero. This experiment suggests that the performance of FO-GPR is much better than the GPR in real applications in terms of both time saving and the accuracy of the learned deformation model. 

\subsubsection{Comparison of online and offline GPR} In this experiment, we fix the Gram matrix unchanged after a while in the rolled towel manipulation task, and compare the performance of the resulting offline model with that of our online learning approach. As shown in Figure~\ref{fig:online_offline}, the error in the feature space $\|\Delta \mathbf x\|_2$ decreases at the beginning of manipulation while using both models for control. However, when the soft object is close to its target configuration, the controller using the offline model cannot output accurate prediction due to the lack of data around the unexplored target state. Thanks to the balance of exploration and exploitation of online FO-GPR, our method updates the deformation model all the time and thus is able to output a relative accurate prediction so that the manipulation process is successful.

\subsubsection{Comparison of FO-GPR and linear model}
We compare our approach to the state-of-the-art online learning method for soft objects' manipulation~\cite{Alarcon:2016:ADM}, which uses a linear model for the deformation function. First, we first through the experiment that the learning rate of the linear model has a great impact on the manipulation performance and needs to be tuned offline for different tasks; while our approach is able to use the same set of parameters for all tasks. Next, we perform both methods on the rolled towel and the peg-in-hole with stretchable fabric tasks, and the results are shown in Figure~\ref{fig:FO_Linear4d1} and~\ref{fig:FO_Linear6d}, respectively. 
To visualize the comparison results, we choose one dimension from the feature vector $\mathbf x$ and plot it. In Figure~\ref{fig:FO_Linear4d1}, we observe that the error of the controller based on the linear model decreases quickly, but the due to the error in other dimensions the controller still outputs a high control velocity and thus vibration starts. The controller needs a long time to accomplish the task. As a contrast, the error of the plotted dimension decreases slower but the controller finishes the task faster because the error of all dimensions declines to zero quickly, thanks to the nonlinear modeling capability of GPR. In the peg-in-hole task, we can observe that the GPR-based controller successfully accomplish the task while the controller based on the linear model fails.

\section{Conclusion and Limitations}
\label{sec:conclusion}
In this paper, we have presented a general approach to automatically servo-control soft objects using a dual-arm robot. We proposed an online GPR model to estimate the deformation function of the manipulated objects, and used low-dimension features to describe the object's configuration. The resulting GPR-based visual servoing system can generate high quality control velocities for the robotic end-effectors and is able to accomplish a set of manipulation tasks robustly. 

For future work, we plan to find a better exploration method to learn a more complicate deformation function involving not only the feature velocity but also the current configuration of the object in the feature space, in order to achieve more challenging tasks like cloth folding.



{\small
\bibliographystyle{IEEEtran}
\bibliography{references}
}

\end{document}